\documentclass[12pt]{iopart}

\usepackage{iopams, bm}
\usepackage{amssymb}
\usepackage{graphicx}
\usepackage{pseudocode}

 \expandafter\let\csname \endcsname\relax
\renewcommand{\thepseudocode}{1}

\begin{document}

\title[AMP for Nonconvex Sparse Regularization]
{Approximate message passing for nonconvex sparse regularization with stability and asymptotic analysis}

\author{Ayaka Sakata$^{1,2}$ and Yingying Xu$^3$}

\address{$^1$The Institute of Statistical Mathematics, Midori-cho, Tachikawa 190-8562, Japan}
\address{$^2$The Graduate University for Advanced Science (SOKENDAI), Hayama-cho, Kanagawa 240-0193, Japan}
\address{$^3$Department of Computer Science, School of Science, Aalto University, P.O. Box 15400, FI-00076 Aalto, Finland}
\ead{$^{1,2}$ayaka@ism.ac.jp,
$^{3}$yingying.xu@aalto.fi}
\vspace{10pt}

\begin{abstract}
We analyse a linear regression problem with nonconvex regularization called smoothly clipped absolute deviation (SCAD) under an overcomplete Gaussian basis for Gaussian random data. We propose an approximate message passing (AMP) algorithm considering nonconvex regularization, namely SCAD-AMP, and analytically show that the stability condition corresponds to the de Almeida--Thouless condition in spin glass literature. Through asymptotic analysis, we show the correspondence between the density evolution of SCAD-AMP
and the replica symmetric solution. Numerical experiments confirm that for a sufficiently large system size, SCAD-AMP achieves the optimal performance predicted by the replica method. Through replica analysis, a phase transition between replica symmetric (RS) and replica symmetry breaking (RSB) region is found in the parameter space of SCAD. The appearance of the RS region for a nonconvex penalty is a significant advantage that indicates the region of smooth landscape of the optimization problem.
Furthermore, we analytically show that the statistical representation performance of the SCAD penalty is better than that of $\ell_1$-based methods, and the minimum representation error under RS assumption is obtained at the edge of the RS/RSB phase.
The correspondence between the convergence of the existing coordinate descent algorithm and RS/RSB transition is also indicated.

\end{abstract}

\section{Introduction}
Variable selection is a basic and important problem in statistics, the objective of which is to find parameters that are significant for the description of given data as well as for the prediction of unknown data.
The sparse estimation approach for variable selection in high-dimensional statistical modelling has the advantages of high computational efficiency, stability,
and the ability to draw sampling properties compared with traditional approaches that follow stepwise and subset selection procedures \cite{Breiman1996}. It has been studied intensively in recent decades.
The development of sparse estimation has been accelerated
since the proposal of the least absolute shrinkage and selection operator (LASSO) \cite{Tibshirani1996},
where variable selection is formulated as
a convex problem of the
minimization of the loss function
associated with $\ell_1$ regularization. Although the LASSO has many attractive properties, the shrinkage introduced by $\ell_1$ regularization results in a significant bias toward 0 for regression coefficients.
To resolve this problem, nonconvex penalties have been proposed,
such as the smoothly clipped absolute deviation (SCAD) penalty \cite{SCAD} and the
minimax concave penalty (MCP) \cite{MCP}.
The estimators under these
regularizations have several desirable properties,
including unbiasedness, sparsity, and continuity \cite{SCAD}.
Nonconvex penalties are seemingly difficult to tackle
because of the concerns regarding local minima owing to
a lack of convexity.
Thus, the development of efficient algorithms and verification of their typical performance have not been achieved.
In previous studies, it has been shown that coordinate descent (CD) is an
efficient algorithm for such nonconvex penalties.
A sufficient condition in parameter space for CD to converge to the globally stable state has been proposed in \cite{Breheny2011}. It was derived by satisfying the convexity of the objective function in a local region of the parameter space that contains the sparse solutions;
however, the convergence condition has not been derived mathematically.
Although the optimal potential of the nonconvex sparse penalty remains unclear,
the existence of such a workable region would imply that the theoretical method applied to the convex penalty is partially valid for the nonconvex penalty.
This factor motivates us to theoretically evaluate the performance and develop an algorithm that is theoretically guaranteed to fulfil the potential of the nonconvex sparse penalty.

In this study, we propose an approximate message passing (AMP) algorithm for linear regression with nonconvex regularization.
For comparison with the $\ell_1$ penalty,
we employ the SCAD penalty, where the nonconvexity is controlled by parameters and
$\ell_1$ regularization is contained at a limit of the parameters.
We start with a brief review of the 
SCAD penalty in the context of variable selection in the following subsections.

\subsection{Smoothly clipped absolute deviation}

\begin{figure}
  \centering
  \includegraphics[width=3.3in]{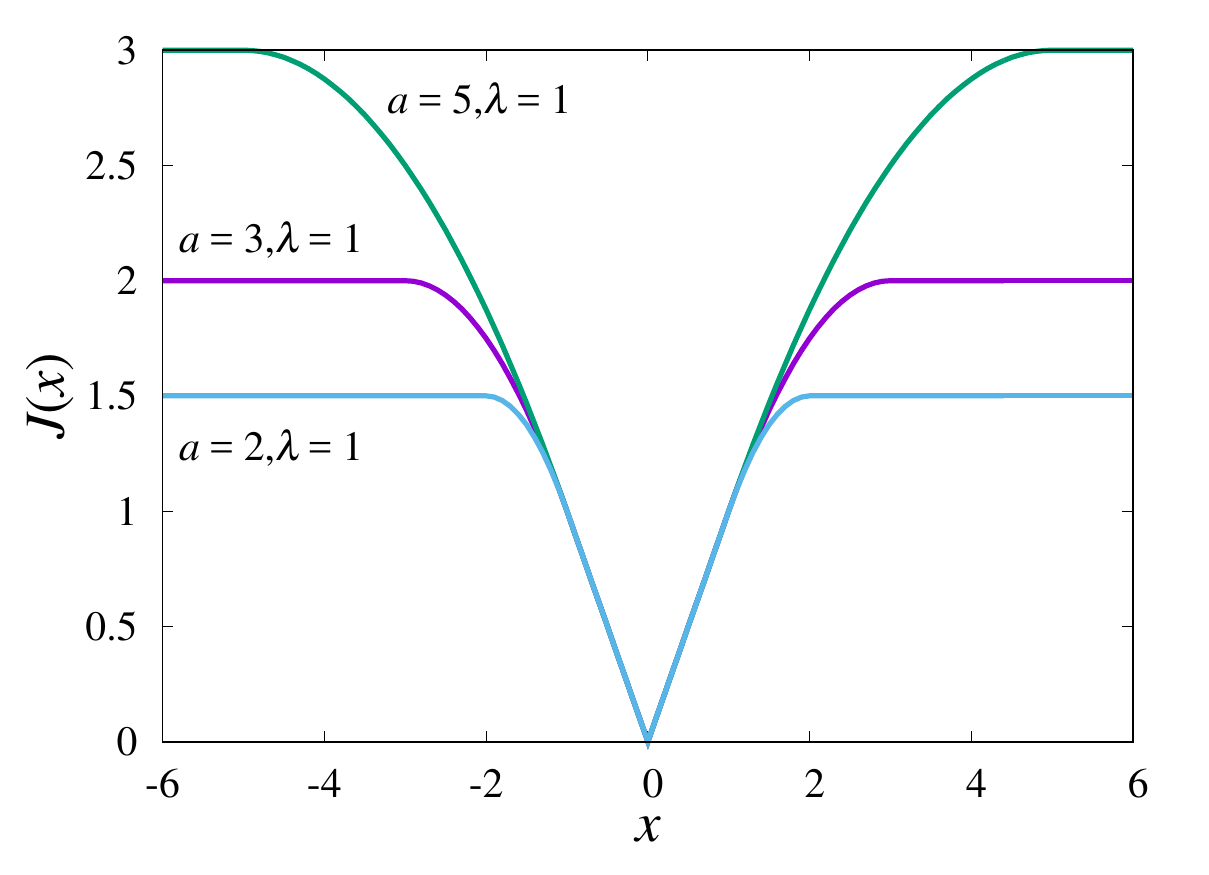}
  \caption{Shapes of SCAD regularizations for $a=2$, 3, and 5 at $\lambda=1$.}
  \label{fig:SCAD_examples}
\end{figure}
Smoothly clipped absolute deviation (SCAD)
is a nonconvex regularization 
defined as
\begin{eqnarray}
J_{\lambda,a}(x)=\left\{\begin{array}{ll}
\lambda|x| & (|x|\leq \lambda) \\
-\displaystyle\frac{x^2-2a\lambda|x|+\lambda^2}{2(a-1)} & (\lambda<|x|\leq a\lambda) \\
\displaystyle\frac{(a+1)\lambda^2}{2} & (|x|>a\lambda)
\end{array}
\right.,
\label{eq:SCAD_def}
\end{eqnarray}
where $\lambda$ and $a(>1)$ are parameters that
control the form of SCAD regularization.
In particular, $a$ contributes to
the nonconvexity of the SCAD function.
\Fref{fig:SCAD_examples} shows
the dependence of SCAD regularization on $a$ at $\lambda=1$.
In the range $[-\lambda,\lambda]$,
SCAD regularization behaves like $\ell_1$ regularization.
Above $a\lambda$ and below $-a\lambda$,
SCAD regularization corresponds to $\ell_0$  regularization
($\ell_0$ norm)
in the sense that the regularization has a constant value.
These $\ell_1$ and $\ell_0$
regions are connected by a quadratic function.
At the limit $a\to\infty$,
the regions $[\lambda,a\lambda]$ and $[-a\lambda,-\lambda]$
become linear with gradients $\lambda$ and $-\lambda$, respectively. Hence,
SCAD regularization is reduced to $\ell_1$ regularization at $a\to\infty$.

\subsection{Overview of variable selection by sparse  regularizations}
\label{sec: regularization}

SCAD regularization is proposed as one of the appropriate 
regularizations for variable selection which provide the following properties to the resulting estimator \cite{SCAD}.
\begin{itemize}
\item Unbiasedness:

To avoid excessive modelling bias,
the resulting estimator is nearly unbiased
when the true unknown parameter is sufficiently large.

\item Sparsity:

The resulting estimator should obey a thresholding rule
that sets small estimated coefficients to zero
and reduces model complexity.

\item Continuity:

The resulting estimator should be continuous with respect to data to avoid instability in model prediction.
\end{itemize}

\begin{figure}
    \centering
    \includegraphics[width=2in]{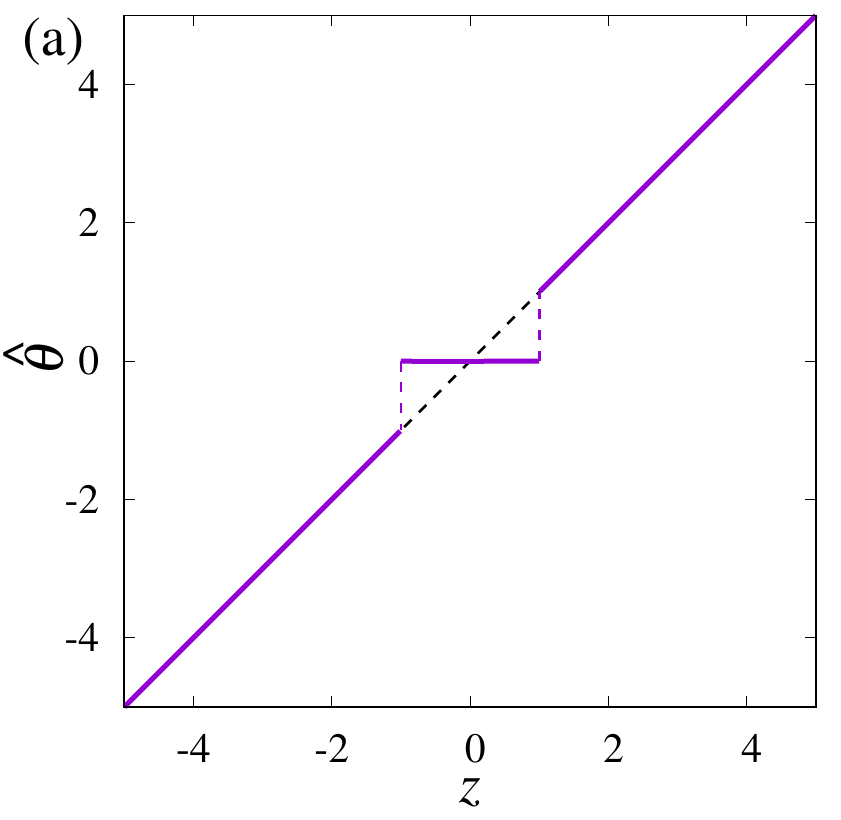}
    \includegraphics[width=2in]{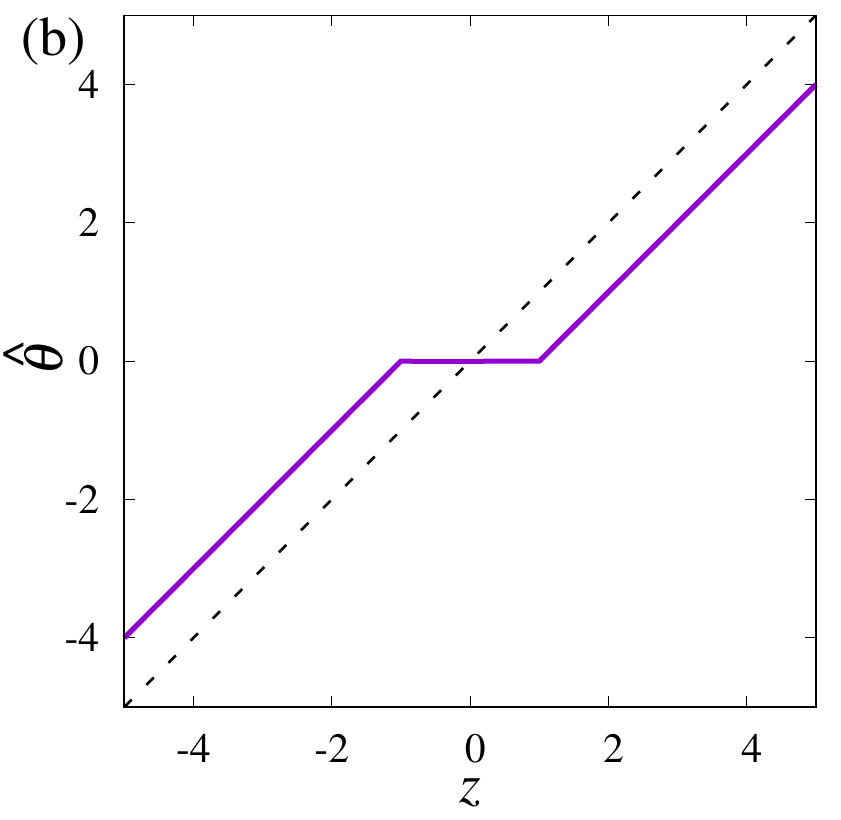}
    \includegraphics[width=2in]{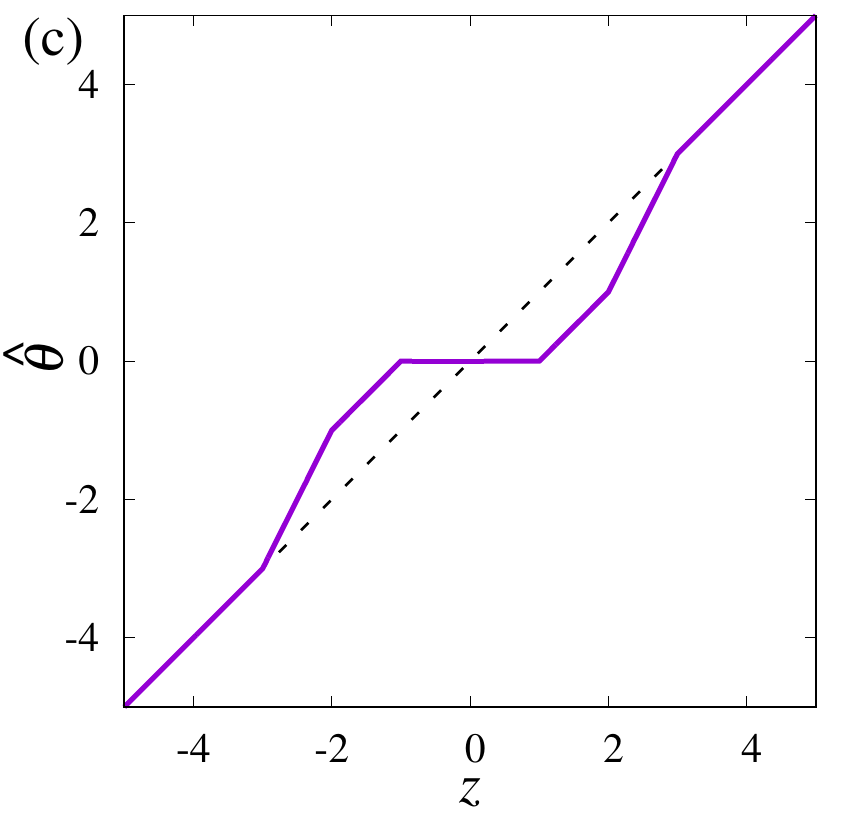}
    \caption{$z$-dependence of the estimator of \eref{eq:example} $\hat{\theta}(z)$ for (a) $\ell_0$, (b) $\ell_1$, and (c) SCAD regularization at $\lambda=1$ and $\sigma_z=1$ for problem \eref{eq:example}. In (c), SCAD parameter is set as $a=3$.
    The diagonal dashed lines represent the least square estimator.}
    \label{fig:estimators}
\end{figure}

For understanding these properties,
let us consider a one-dimensional minimization problem
under regularization $J(\theta)$ defined by
\begin{eqnarray}
\hat{\theta}(z)=\arg\min_\theta\left\{\frac{1}{2\sigma_z^2}(z-\theta)^2+J(\theta)\right\},
\label{eq:example}
\end{eqnarray}
where $z$ is the given data and $\theta$ is the variable to be estimated.
The least square (LS) estimator, which corresponds to the estimate for $J(\theta)$ constantly equals to zero, 
is $\hat{\theta}_{\mathrm{LS}}(z)=z$ for any value of $z$.
The presence of a nonzero regularization function generally 
shifts the actual estimate from the least square one.
As a result, the squared error (representation error) between data $z$ and its fit $\hat{\theta}$ increases.
Therefore, the reduction of the bias of the estimator from LS one is one of the challenges for the penalized regression problems.
\Fref{fig:estimators} shows the behaviour
of the estimator for 
$\ell_0$ ($J(\theta)=\lambda|\theta|_0$), $\ell_1$ ($J(\theta)=\lambda|\theta|_1$), and 
SCAD ($J(\theta)=J_{\lambda,a}(\theta))$ regularizations where parameters are set as $\lambda=1, a=3$ and $\sigma_z=1$.
The bias of the estimator is depicted as the difference between the least square estimator (diagonal dashed lines) and the actual estimator (solid lines).
As shown in \Fref{fig:estimators}(a), the estimator under $\ell_0$ regularization ($\ell_0$-estimator) shows a discontinuous jump at $z=\lambda$
from zero to nonzero value which corresponds to the LS estimator,
and not continuous with respect to $z$.
The discontinuous estimator is sensitive to the fluctuating data which is not desirable for variable selection.
The estimator under $\ell_1$ regularization ($\ell_1$-estimator) shown in \Fref{fig:estimators} (b) continuously changes from zero to nonzero value at $z=\lambda$, but is shrunk from LS estimators for any $z$, and this indicates the increase in the representation error.
The estimator under SCAD regularization (SCAD estimator), shown in \Fref{fig:estimators} (c), behaves as $\ell_1$ and $\ell_0$ estimator at $\lambda< z\leq2\lambda$, and $z>a\lambda$, respectively.
The estimate at $z>a\lambda$ is not biased despite the existence of regularization, and hence
this region contributes to the unbiasedness of the estimates under SCAD regularization.
At $2\lambda< z\leq a\lambda$, the SCAD estimator transits linearly between the $\ell_1$ and $\ell_0$ estimators.
This continuous change is provided by the quadratic function of SCAD regularization \eref{eq:SCAD_def} at $\lambda<|\theta|\leq  a\lambda$.
In this manner, SCAD regularization \eref{eq:SCAD_def} is designed to simultaneously achieve 
unbiasedness, continuity, and sparsity.

\begin{table}
\centering
\begin{tabular}{|c||c|c|c|}\hline
Regularization & Unbiasedness & Sparsity & Continuity \\\hline\hline
$\ell_q$ $(q>1)$ & $\times$ & $\times$ & $\surd$ \\\hline
$\ell_1$ & $\times$ & $\surd$ & $\surd$ \\\hline
$\ell_q$ $(q<1)$ & $\surd$ & $\surd$ & $\times$ \\\hline
SCAD & $\surd$ & $\surd$ & $\surd$ \\\hline
\end{tabular}
\caption{Properties of estimator under representative regularizations for parameter selection problem.}
\label{table:penalties}
\end{table}

\Tref{table:penalties} summarizes the properties of the estimator for parameter selection problem under representative regularizations.
The regression problem with
$\ell_q$ regularization is called bridge regression \cite{Frank1993}.
A well-known example for $q>1$ is ridge regression
associated with the $\ell_2$ penalty \cite{ridge}.
The resulting estimator under $\ell_q~(q>1)$
has continuity, but it is not sparse;
hence, $\ell_q~(q>1)$ regularization is not
appropriate for variable selection.
The $\ell_q~(q<1)$ penalty gives a sparse and unbiased estimator,
but the estimator is discontinuous, which is the same as the $\ell_0$ case.

\subsection{Our main contributions}
In this paper, we focus on the linear regression problem in high-dimensional settings where target data $\bm{y}\in\mathbb{R}^M$
is approximated by using an overcomplete predictor matrix or basis matrix $\bm{A}\in\mathbb{R}^{M\times N}$ ($M<N$) as $\bm{y}\sim\bm{Ax}$.
The regression coefficient $\bm{x}\in\mathbb{R}^N$ is to be estimated under a nonconvex sparse penalty.
Among the given overcomplete basis matrix, a small combination of basis vectors is selected corresponding to the regularization parameters, which compactly and accurately represent the target data. 
The sparse representation under the overcomplete basis
is an appropriate problem to check the variable selection ability of
sparse regularization.

The organization and main contributions of this paper are summarized below:

\begin{itemize}

\item We propose an efficient algorithm based on message passing,
namely SCAD-AMP, for a SCAD regularised linear regression problem (Section \ref{sec:SCAD-AMP}), and we analytically show
its local stability condition
in Section \ref{sec: microscopic stability}.

\item Through asymptotic analysis, the density evolution of the message passing algorithm is shown in Section \ref{sec:density_evolution}. 
In addition, the validity of the asymptotic analysis is confirmed by numerical simulation.

\item Based on the replica analysis, a phase transition between the replica symmetric (RS) and replica symmetry breaking (RSB) regions is found in the parameter space of the SCAD  regularization
(Section \ref{sec:phasetransition_error}).
In other words, the RS phase exists when the nonconvexity of the regularization is appropriately controlled.
We also confirm that the conventional parameter setting $a=3.7$ corresponds to the RS phase for sufficiently large $\lambda$.

\item We show that the stability of the RS solution corresponds to that of AMP's fixed points. This means that the local stability of AMP is mathematically guaranteed in the RS phase.
This correspondence holds regardless of the type of regularization. Hence, the result is valid for other regularizations where the RS/RSB transition appears (section \ref{sec:AT_RS}).

\item We evaluate the statistically optimal 
value of the error between $\bm{y}$ and $\bm{Ax}$ for the SCAD regularized linear regression problem.
The error has a decreasing tendency as parameter $a$ decreases and approaches the RS/RSB boundary,
and it is the lowest at the edge of the RS/RSB phase.
In addition, we analyze the parameter dependence of the density of the nonzero components in $\bm{x}$ in the RS phase. The sparsity is nearly controlled by $\lambda$, but as $a$ decreases, the number of nonzero components slightly increases.
$\ell_1$ regularization corresponds to $a\to\infty$ of SCAD regularization; hence,
SCAD typically provides a more accurate and sparser expression compared with $\ell_1$ in the RS phase (section \ref{sec:phasetransition_error}).

\item It is numerically shown that the RS/RSB transition point corresponds to the limit at which the coordinate descent algorithm reaches the globally stable solution (Section \ref{sec:Coordinate_Descent}).

\item Our analysis shows that
the transient region of SCAD estimates between the $\ell_1$-type and the $\ell_0$-type contributes to the occurrence of the RSB transition (Section \ref{sec:AT_RS}).

\end{itemize}

\section{Problem settings}
\label{sec:problem_setting}

In this paper, we analyse a linear regression problem with
SCAD regularization,
which is formulated as
\begin{eqnarray}
  \min_{\bm{x}}\frac{1}{2}||\bm{y}-\bm{Ax}||_2^2+J_{\lambda,a}(\bm{x}),
\label{eq:SCAD_LR}
\end{eqnarray}
where $\bm{y}\in\mathbb{R}^M$ and
$\bm{A}\in\mathbb{R}^{M\times N}$
are the given data and the predictor matrix or basis matrix,
respectively.
Here, we consider the case in which the generative model of $\bm{y}$
does not contain a true sparse expression;
hence, \eref{eq:SCAD_LR} corresponds to a
compression problem of the given data $\bm{y}$ under the basis $\bm{A}$,
rather than the reconstruction of the true signal.
We assume that each component of $\bm{y}$ and $\bm{A}$
is independently and identically distributed (i.i.d) according to
Gaussian distributions.
More specifically,
their joint distribution is given by
\begin{eqnarray}
&P_{\bm{y},\bm{A}}(\bm{y},\bm{A})=P_{\bm{y}}(\bm{y})P_{\bm{A}}(\bm{A})\label{eq:P_yA}\\
&P_{\bm{y}}(\bm{y})=\prod_{\mu=1}^M\frac{1}{\sqrt{2\pi\sigma_y^2}}\exp\left(-\frac{y_\mu^2}{2\sigma_y^2}\right)\\
&P_{\bm{A}}(\bm{A})=\prod_{\mu=1}^M\prod_{i=1}^N\sqrt{\frac{M}{2\pi}}\exp\left(-\frac{MA_{\mu i}^2}{2}\right).
\end{eqnarray}
In conventional analysis by statistical mechanics for linear systems, such as compressed sensing problems, the elements of the matrix $\bm{A}$ are often assumed with zero mean and variance $1/N$. We set the variance to $1/M$ to match the conventional assumptions in the statistics literature, where $\sum_{\mu}A_{\mu i}^{2}=1$ holds for all $i$ asymptotically or for average of many $\bm{A}$ sets.
The coefficient $1\slash 2$ of \eref{eq:SCAD_LR}
is introduced for mathematical convenience, and
we denote the function to be minimized divided by $M$ as
\begin{eqnarray}
e(\bm{x}|\bm{y},\bm{A})\equiv \frac{1}{M}\left\{\frac{1}{2}||\bm{y}-\bm{Ax}||_2^2+J_{\lambda,a}(\bm{x})\right\},
\label{eq:energy}
\end{eqnarray}
which corresponds to the energy density.
When the regularization is given by
$\ell_1$ norm,
the problem is known as the LASSO \cite{Tibshirani1996}.
SCAD regularization induces zero components into the estimated variable $\bm{x}$
corresponding to parameters $\lambda$ and $a$.
We set $\lambda$ and $a$ as $O(1)$
regardless of the system size.

Further, we introduce the posterior distribution of the estimate
with a parameter $\beta$:
\begin{eqnarray}
P_{\beta}(\bm{x}|\bm{y},\bm{A})=\exp\left\{-\beta Me(\bm{x}|\bm{y},\bm{A})-\ln Z_\beta(\bm{y},\bm{A})\right\},
\label{eq:prob}
\end{eqnarray}
where $Z_\beta(\bm{y},\bm{A})$ is the normalization constant.
The limit $\beta\to\infty$ leads to the uniform distribution over the minimizers of \eref{eq:SCAD_LR}.
The estimate of the solution of \eref{eq:SCAD_LR} under a fixed set of $\{\bm{y},\bm{A}\}$,
denoted by $\hat{\bm{x}}(\bm{y},\bm{A})$, is given by
\begin{eqnarray}
\hat{\bm{x}}(\bm{y},\bm{A})=\lim_{\beta\to\infty}\langle\bm{x}\rangle_{\beta},
\label{x_expectation}
\end{eqnarray}
where $\langle\cdot\rangle_\beta$
denotes the expectation according to \eref{eq:prob} at $\beta$.

We focus on the overcomplete basis,
where the number of column vectors is greater than
the dimension of the data, i.e. $N>M$, with the compression ratio $\alpha=M\slash N~(< 1)$.
We do not impose orthogonality on the basis vectors.
In general, the overcomplete basis gives an infinitely large number of solutions,
but by adding the cost functions that promote sparsity,
a more informative representation than that under the orthogonal basis is potentially obtained.
The sparse representation under the overcomplete basis
is an appropriate problem to check the variable selection ability of the
sparse regularization.

\section{Approximate message passing algorithm}
\label{sec:AMP}
The exact computation of the expectation in (\ref{x_expectation}) is intractable since it requires exponential time \cite{Natarajan1995}. We develop an approximate algorithm for (\ref{x_expectation}) following the framework of belief propagation (BP) or message passing \cite{Mezard1987, Mezard2009, MacKay1999, Kabashima1998}.
BP has been developed for problems with sparse regularizations, such as compressed sensing
with linear measurements \cite{Donoho2009}
and LASSO \cite{DMM_IEEE,Bayati-Montanari},
exhibiting high reconstruction accuracy and computational efficiency.
To incorporate the nonconvex penalty, we employ a variant of BP
known as generalized approximate message passing (GAMP) \cite{Rangan2011, Kabashima2004GAMP}.
The details of the derivation of the SCAD-AMP algorithm are described in \ref{app:derivation_of_SCAD_AMP}. In the following, we first briefly summarise AMP in Sec. \ref{sec:general_BP} and Sec. \ref{sec:general_AMP} based on earlier studies \cite{Krzakala2012,Xu2014}, and then derive a specific form for SCAD in Sec. \ref{sec:SCAD-AMP},
which is one of our contributions.
Its asymptotic analysis is shown after the derivation in Sec. \ref{sec:density_evolution}. Stability analysis, which is an essential and a common concern of algorithms developed for nonconvex regularizations, is discussed at the end of the section in Sec. \ref{sec: microscopic stability}.

\subsection{General form of belief propagation}
\label{sec:general_BP}

\begin{figure}
    \centering
    \includegraphics[width=5in]{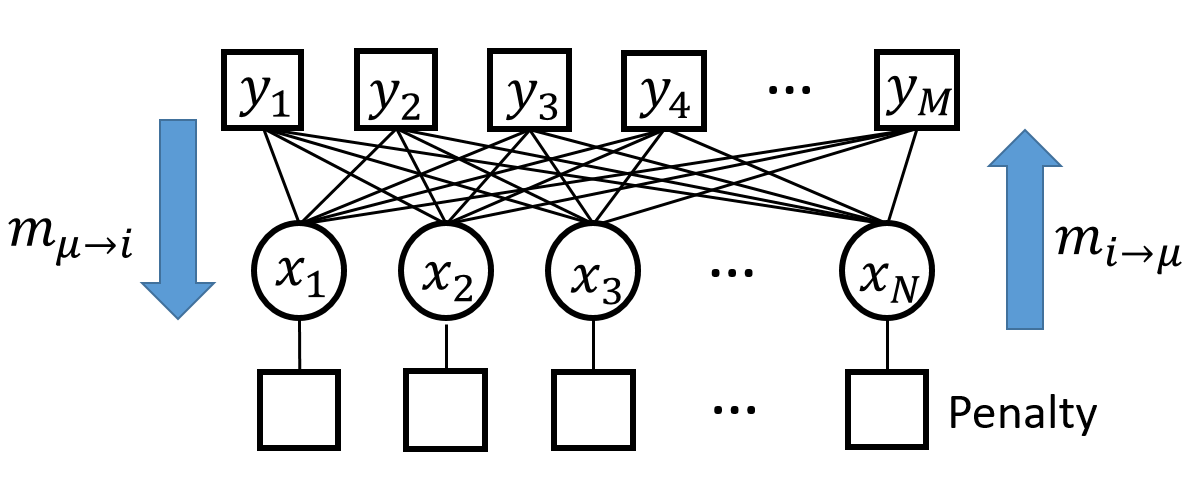}
    \caption{Graphical representation of belief propagation.}
    \label{fig:dense_graph}
\end{figure}

In message passing, 
the probability system can be 
graphically denoted by two types of nodes and connects them by an edge when they are related \cite{Rangan2011,Krzakala2012}. The conditional probability of $y_{\mu}$ depends on all of $x_1, x_2, \cdots, x_{N}$, implying that the posterior distribution \eref{eq:prob} can be expressed as a (dense) complete bipartite graph, as shown in Figure \ref{fig:dense_graph}. Let us define a general constraint/penalty/prior distribution for the variable nodes as $P_r\left(\bm{x}\right)$, and a probability distribution for the output function notes as $P\left(\bm{y}|\bm{Ax}\right)$.
The canonical BP equations for the probability $P(\bm{x}|\bm{A},\bm{y})$ are
generally expressed in terms of $2M\times N$ messages,
$m_{i\rightarrow \mu}\left(x_i\right)$ and $m_{\mu\rightarrow i}\left(x_i\right) (i=1,2,\cdots,N~{\rm and}~ \mu=1,2,\cdots,M)$,
which represent the probability distribution functions that carry posterior information and output information, respectively.
The messages are written as
\begin{eqnarray}
m_{\mu\rightarrow i}\left(x_i\right)
=\frac{1}{Z_{\mu\rightarrow i}}\int\prod_{j \neq i}\textrm{d} x_j P\left(y_{\mu}|\left(\bm{A}\bm{x}\right)_{\mu}\right) \prod_{j \neq i} m_{j\rightarrow \mu}\left(x_j\right), \label{m_mu_i}\\
m_{i\rightarrow \mu}\left(x_i\right)
=\frac{1}{Z_{i\rightarrow \mu}} P_r\left(x_i\right)\prod_{\gamma \neq \mu}m_{\gamma\rightarrow i}\left(x_i\right),\label{m_i_mu}
\end{eqnarray}
where $Z_{\mu\rightarrow i}$ and $Z_{i\rightarrow \mu}$ are normalization constants that satisfy
$\int\textrm{d} x_i m_{\mu\rightarrow i}(x_{i})=\int\textrm{d} x_i m_{i\rightarrow \mu}(x_{i})=1$.
The means and variances of $x_i$ under the posterior information message distributions are defined by  
\begin{eqnarray}
a_{i\rightarrow\mu}\equiv \int\textrm{d}x_{i}x_{i}m_{i\rightarrow \mu}\left(x_i\right), \label{a_imu}\\
\nu_{i\rightarrow\mu}\equiv \int\textrm{d}x_{i}x_{i}^{2}m_{i\rightarrow \mu}\left(x_i\right)-a_{i\rightarrow\mu}^2 \label{nu_imu}.
\end{eqnarray}
We also define
\begin{eqnarray}
\omega_{\mu}&\equiv \sum_{i}A_{\mu i}a_{i\rightarrow\mu}\label{omega_df},\\ 
V_{\mu}&\equiv \sum_{i}A_{\mu i}^2 \nu_{i\rightarrow\mu}\label{Vmu_df},
\end{eqnarray}
for notational convenience.
Using (\ref{m_mu_i}), we evaluate the approximation of marginal distributions (beliefs) as
\begin{eqnarray}
m_{i}\left (x_i \right )
=\frac{1}{Z_i} P_r(x_i)\prod_{\mu=1}^M m_{\mu \rightarrow i} \left(x_i\right), \label{m_i}
\end{eqnarray}
where $Z_i$ is a normalization constant for $\int {\rm d} x_i m_i\left (x_i \right )=1$.
We denote the means and variances of the beliefs as $a_i$ and $\nu_i$, respectively, which are given by
\begin{eqnarray}
a_{i}&\equiv \int\textrm{d}x_{i}x_{i}m_{i}\left(x_i\right),\\
\nu_{i}&\equiv \int\textrm{d}x_{i}x_{i}^{2}m_{i}\left(x_i\right)-a_{i}^2.
\end{eqnarray}
The mean $\bm{a}=(a_i)$ represents the approximation of the posterior mean $\hat{\bm{x}}(\bm{y},\bm{A})$, which is the output of the algorithm.

\subsection{General form of AMP}
\label{sec:general_AMP}

After the calculations shown in \ref{app:derivation_of_SCAD_AMP} under assumptions
on the predictor matrix that the correlation between the components are negligible and $A_{\mu i}\sim O(N^{-1\slash 2})$,
we obtain the marginal distribution as 
\begin{equation}
m_i(x)={\cal M}(x;\Sigma^2,R_i),
\label{eq:marginal_dist}
\end{equation}
where 
\begin{equation}
    {\cal M}(x;\Sigma^2,R)=\frac{1}{\hat{Z}(\Sigma^2, R)}P_r(x){\frac{1}{\sqrt{2\pi\Sigma^2}}}\exp\left(-\frac{(x-R)^2}{2\Sigma^2}\right),
    \label{M_dist}
\end{equation}
and $\hat{Z}(\Sigma^2,R)$ is the normalization constant.
\Eref{eq:marginal_dist} means that the product of messages $\prod_\mu m_{\mu\to i}(x_i)$
is approximated as a Gaussian distribution of mean $R_i$ and variance $\Sigma^2$ given by
\begin{eqnarray}
\Sigma^{2}=\left( \frac{1}{M}\sum_{\mu}(g_{\rm out}^{\prime})_{\mu} \right)^{-1}, \label{eq:Sigma}\\
R_i=a_{i}+\left( \sum_{\mu}(g_{\rm out})_{\mu}A_{\mu i}\right)\Sigma^{2},\label{eq:R}
\end{eqnarray}
where
\begin{eqnarray}
(g_{\rm out})_{\mu} &\equiv& \frac{\partial }{\partial \omega_\mu}
\log \left (\int {\rm d} u_\mu P(y_\mu|u_\mu)\exp \left (-\frac{(u_\mu-\omega_\mu)^2}{2V} \right ) \right )
\label{eq:g_out_GAMP} \\
(g^\prime_{\rm out})_{\mu} &\equiv & -\frac{\partial^2 }{\partial \omega_\mu^2}
\log \left (\int {\rm d} u_\mu P(y_\mu|u_\mu)\exp \left (-\frac{(u_\mu-\omega_\mu)^2}{2V} \right ) \right ),
\label{eq:g_out_p_GAMP}
\end{eqnarray}
and
\begin{eqnarray}
V&=\frac{1}{M}\sum_{i=1}^Nv_i^2\\
\omega_{\mu}&= \sum_{i} A_{\mu i}a_i -(g_{\rm out})_{\mu}V. \label{eq:omega_general}
\end{eqnarray}
We note that $a_i$ in equation (\ref{eq:R}) and $(g_{\rm out})_{\mu}V$ in equation (\ref{eq:omega_general}) correspond to
the {\em Onsager reaction term} in the spin glass literature  \cite{Thouless1977,Shiino1992}.
These terms effectively cancel out the self-feedback effects, thereby stabilising the convergence of GAMP.

We define the following function associated with the distribution ${{\cal M}}(x;\Sigma^2, R)$,
\begin{eqnarray}
f_{a}(\Sigma^2, R)\equiv \int\textrm{d}xx{\cal{M}}(x;\Sigma^2, R),\label{fa_general}\\
f_{c}(\Sigma^2, R)\equiv \int\textrm{d}xx^{2}{\cal{M}}(x;\Sigma^2, R)-f_{a}^{2}(\Sigma^2, R)\label{fc_general}.
\end{eqnarray}
Hence $a_i$ and $v_i$ are given as follows.
\begin{eqnarray}
a_i&=f_a(\Sigma^2,R_i)\\
v_i&=f_c(\Sigma^2,R_i)
\end{eqnarray}
Note that $f_a$ and $f_c$ satisfy the following relationships
\begin{eqnarray}
f_{a}(\Sigma^2, R)=R+\Sigma^{2}\frac{\partial}{\partial R}\textrm{log}\hat{Z}(\Sigma^2, R),\\
f_{c}(\Sigma^2, R)=\Sigma^{2}\frac{\partial}{\partial R}f_a(\Sigma^2, R).
\end{eqnarray}
The form of $f_a$ and $f_c$ depends only on the distribution $P_r(x)$. For a general penalty distribution of $x$, the functions $f_a$ and $f_c$ are computed by numerical integration over $x$. In special cases, such as the Bernoulli Gaussian or $l_1$ penalty, these functions are easily computed analytically \cite{Xu2014}, as in the case of the SCAD penalty. We give the specific analytic form of $f_a, f_c$ for the SCAD penalty in the following section.

\subsection{SCAD-AMP for linear regression}
\label{sec:SCAD-AMP}

\begin{figure}
\renewcommand{\thepseudocode}{\arabic{pseudocode}}
\setcounter{pseudocode}{0}
\begin{pseudocode}[ruled]
{Approximate Message Passing for SCAD}
{\bm{a}^*, \bm{\nu}^*, \bm{\omega}^*}

1)\ \mbox{\bf Initialization}:\\
 \hspace{15pt}\bm{a}\textrm{ seed}: \hspace{65pt} \mathbf{a}_{0} \GETS \mathbf{a}^*\\
 \hspace{15pt}\bm{\nu}\textrm{ seed}: \hspace{65pt} \bm{\nu}_0 \GETS \bm{\nu}^* \\
 \hspace{15pt}\bm{\omega}\textrm{ seed}: \hspace{65pt} \bm{\omega}_0 \GETS \bm{\omega}^* \\
 \hspace{15pt}\textrm{Counter}: \hspace{62pt} t\GETS 0\\

2)\ \mbox{\bf Counter increase}: \hspace{20pt}t \GETS t+1\\

3)\ \mbox{\bf Mean of variances of posterior information message distributions}:\\
 \hspace{30pt}{V}^{(t)}\GETS  M^{-1}(\textrm{sum} (\bm{\nu}^{(t-1)}))\\

4)\ \mbox{\bf Self-feedback cancellation}:\\
 \hspace{30pt}
 \bm{\omega}^{(t)} \GETS \bm{A}\bm{a}^{(t-1)}-\frac{V^{(t)}}{V^{(t)}+1}(\bm{y}-\bm{\omega}^{(t-1)})\\

5)\ \mbox{\bf Average of output information message distributions}:\\
 \hspace{30pt}(\mathbf{R})^{(t)}\GETS \bm{a}^{(t-1)}+\bm{A}(\bm{y}-\bm{\omega}^{(t)})\\

6)\ \mbox{\bf Posterior mean}:\\
 \hspace{30pt}\bm{a}^{(t)}\GETS f_a (({V}^{(t)}+1)\bm{1}, \mathbf{R}^{(t)})\\

7)\ \mbox{\bf Posterior variance}:\\
 \hspace{30pt}\bm{\nu}^{(t)}\GETS f_c (({V}^{(t)}+1)\bm{1}, \mathbf{R}^{(t)})\\

8)\ \mbox{\bf Iteration}: \mbox{Repeat from step 2 until convergence.}
\end{pseudocode}
\protect
\caption{Pseudocode of SCAD-AMP with the assumption that matrix $\bm{A}$ contains i.i.d random variables with zero mean and variance $1/M$.
For the linear regression problem, the variance vector of output information message distributions is $\mathbf{V}+\bm{1}$. Functions $f_a$ and $f_c$ for SCAD are (\ref{fa_SCAD}) and (\ref{fc_SCAD}), respectively.
The convergent vectors of $\bm{a}^{(t)}$, $\bm{\nu}^{(t)}$,
and $\bm{\omega}^{(t)}$ obtained in the previous loop are denoted by $\bm{a}^*$, $\bm{\nu}^*$, and $\bm{\omega}^*$,respectively.
$\bm{1}$ is the $N$-dimensional vector whose entries are all unity. }
\label{proposedalgorithm}
\end{figure}

Approximate message passing for smoothly clipped absolute deviation (SCAD-AMP) is derived from the general form by 
specifying the output channel and penalty distribution.
Figure \ref{proposedalgorithm} shows the pseudocode of SCAD-AMP.
First, let us consider the output channel for the linear regression problem
\begin{equation}
P\left(y_{\mu}|u_{\mu}\right)=\exp\left\{-\frac{\beta}{2}(y_\mu-u_\mu)^2\right\},
\label{LR_output}
\end{equation}
where we focus our analysis on the $\beta \rightarrow \infty$ case. 
We rescale $\tilde{V}=\beta V$ and $\tilde{\Sigma}^2=\beta \Sigma^2$ such that $\tilde{V}$ and $\tilde{\Sigma}^2$ are $O(1)$ in the $\beta \rightarrow \infty$ limit.
Inserting \eref{LR_output} into (\ref{eq:g_out_GAMP}) and \eref{eq:g_out_p_GAMP}
gives $(g_{\rm out})_{\mu}$ and $(g_{\rm out}^{\prime})_{\mu}$ as 
\begin{eqnarray}
(g_{\rm out})_{\mu}=\beta \frac{y_{\mu}-\mathbf{\omega}_{\mu}}{\tilde{V}+1}\label{g_out_SCAD},\\
(g_{\rm out}^{\prime})_{\mu}=\beta\frac{1}{\tilde{V}+1},\label{g_out_p_SCAD}
\end{eqnarray}
respectively. 
Inserting (\ref{g_out_SCAD}) into \eref{eq:omega_general} gives
\begin{equation}
\omega_{\mu}= \sum_{i} A_{\mu i}a_i -\frac{\tilde{V}}{\tilde{V}+1}(y_{\mu}-\omega_{\mu}),
\end{equation}
and inserting (\ref{g_out_p_SCAD}) into \eref{eq:Sigma}
gives
\begin{equation}
\tilde{\Sigma}^2=\tilde{V}+1.
\label{Sigma_SCAD}
\end{equation}

\begin{figure}
\centering
\includegraphics[width=3in]{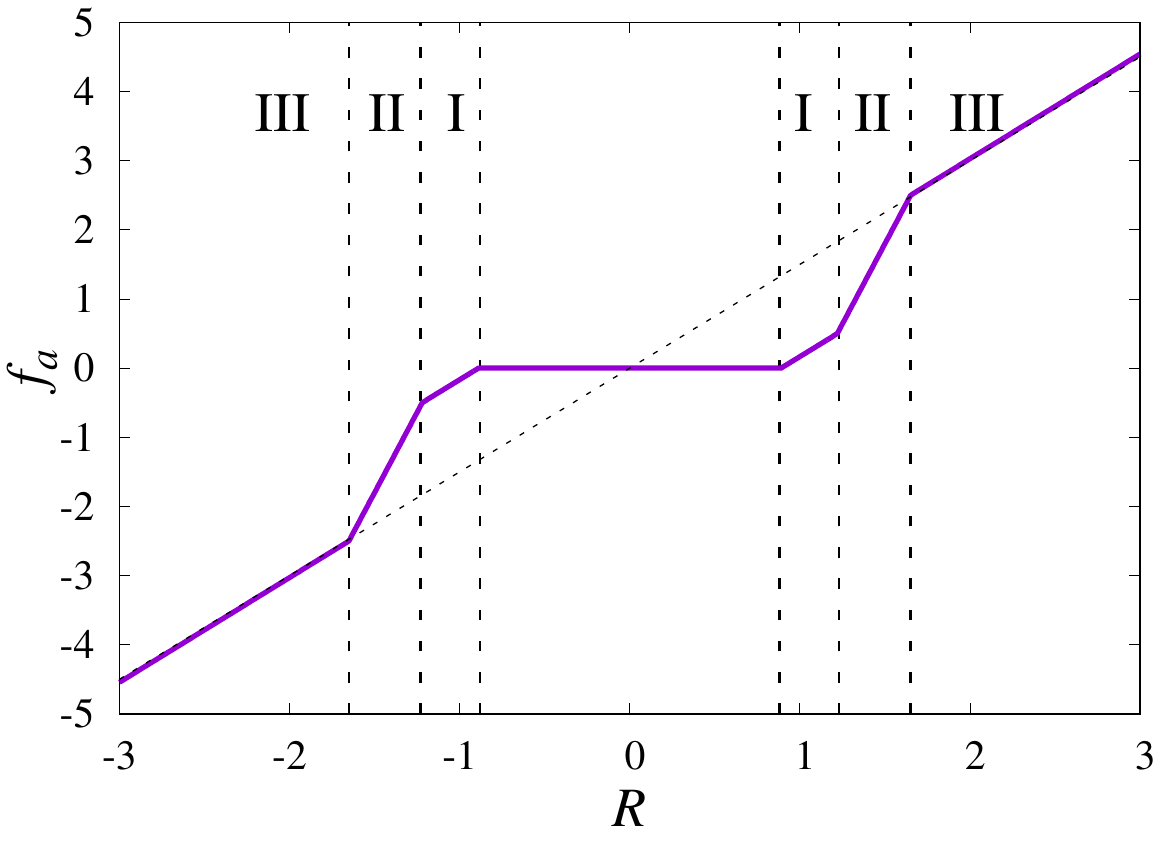}
\caption{Example of $f_a$
as a function of $R$.
The dashed diagonal line
represents the behaviour of the
estimate when the regularization term does not exist.}
\label{fig:SCAD_single_body}
\end{figure}

Next, on the prior distribution side, by inserting SCAD penalty distribution
\begin{equation}
P_r(x)=e^{-\beta J_{\lambda,a}(x)},
\end{equation}
into \eref{M_dist}, in the limit of $\beta\rightarrow\infty$, we obtain
\begin{eqnarray}
\nonumber
\hat{Z}(\Sigma^{2}, R)&=&\int_{-\infty}^{\infty}\sqrt{\frac{\beta}{2\pi\tilde{\Sigma}^2}}\exp\left\{-\beta J_{\lambda, a}(x) - \beta\frac{(x-R)^2}{2\tilde{\Sigma}^2}\right\}dx\\
&=&\sqrt{\frac{\beta}{2\pi\tilde{\Sigma}^2}}\exp\left\{ -\beta\phi^*(\tilde{\Sigma}^2,R)\right\},
\end{eqnarray}
where
\begin{eqnarray}
\phi^*(\tilde{\Sigma}^2,R)=\min_x \phi(x;\tilde{\Sigma}^2,R),\\
\phi(x;\tilde{\Sigma}^2,R)=J_{\lambda, a} (x)+\frac{(x-R)^2}{2\tilde{\Sigma}^2},
\end{eqnarray}
and $R$ is given by inserting \eref{g_out_SCAD} and \eref{Sigma_SCAD} into \eref{eq:R}, which gives
\begin{equation}
R_{i}=\sum_{\mu}A_{\mu i}(y_{\mu}-\omega_{\mu})+a_{i}.\label{R_SCAD}
\end{equation}
At $\beta\to\infty$,
the function $f_a$ corresponds to the minimizer of $\phi$ as
\begin{eqnarray}
f_a(\Sigma^2,R)=\arg\min_x \phi(x;\tilde{\Sigma}^2,R).
\label{eq:f_a_beta_infty}
\end{eqnarray}
By solving the minimization problem, we obtain
\begin{eqnarray}
{f}_a(\tilde{\Sigma}^2, R)=\tilde{\Sigma}^2\left(\frac{R}{\tilde{\Sigma}^2}-\lambda\textrm{sign}\left(\frac{R}{\tilde{\Sigma}^2}\right)\right)\cdot \textrm{I}\nonumber\\
       \hspace{2cm}          +\frac{1}{\frac{1}{\tilde{\Sigma}^2}-\frac{1}{a-1}}\left(\frac{R}{\tilde{\Sigma}^2}-\frac{a}{a-1}\lambda\textrm{sign}\left(\frac{R}{\tilde{\Sigma}^2}\right)\right)\cdot \textrm{I\hspace{-.1em}I}+R\cdot \textrm{I\hspace{-.1em}I\hspace{-.1em}I},\label{fa_SCAD}\\
{f}_c(\tilde{\Sigma}^2, R)=\tilde{\Sigma}^2\cdot\textrm{I}+\frac{1}{\frac{1}{\tilde{\Sigma}^2}-\frac{1}{a-1}}\cdot\textrm{I\hspace{-.1em}I}+\tilde{\Sigma}^2\cdot\textrm{I\hspace{-.1em}I\hspace{-.1em}I},\label{fc_SCAD}
\end{eqnarray}
where
\begin{eqnarray}
\textrm{I}=\Theta\left(\left|R\right| >\lambda\tilde{{\Sigma}}^2\right)\Theta\left(\left|R\right| \leq \lambda (1+\tilde{\Sigma}^2)\right),\label{eq:region1}\\
\textrm{I\hspace{-.1em}I}=\Theta\left(\left|R\right| >\lambda(1+\tilde{\Sigma}^2)\right)\Theta\left(\left|R\right| \leq a\lambda\right),\label{eq:region2}\\
\textrm{I\hspace{-.1em}I\hspace{-.1em}I}=\Theta\left(\left|R\right| > a\lambda\right),\label{eq:region3}
\end{eqnarray}
and $\Theta(x)$ is the step function which takes value 1 when the condition $x$ is satisfied, 0 otherwise.
Here, we have replaced ${f}_a({\Sigma}^2, R)$ and ${f}_c({\Sigma}^2, R)$
with ${f}_a(\tilde{\Sigma}^2, R)$ and
${f}_c(\tilde{\Sigma}^2, R)$,
respectively, since there is no $\beta$ dependence in the final form of (\ref{fa_SCAD}) and (\ref{fc_SCAD}). \Fref{fig:SCAD_single_body} shows an example of $f_a$ as a function of $R$, where regions $\mathrm{I}$, $\mathrm{I\!I}$, and $\mathrm{I\!I\!I}$ represent the region where constraints \eref{eq:region1}, \eref{eq:region2}, and \eref{eq:region3} are satisfied, respectively. This figure is equivalent to \Fref{fig:estimators} (c) with the correspondence of $\sigma_z^2$ and $z$ with $\Sigma_i^2$ and $R$, respectively.
We will discuss the contribution of region $\mathrm{I\!I}$ 
to the instability of SCAD-AMP in connection with the replica method in Section \ref{sec:replica}.

The quantities we are focusing on are calculated at AMP's fixed point.
For example,
we calculate the ratio of the nonzero components to the data dimension, called sparsity, as
\begin{eqnarray}
\frac{\rho}{\alpha} =\frac{1}{M}\sum_{i=1}^N|\hat{x}_i|_0
\end{eqnarray}
at AMP's fixed point.
\Fref{fig:delta_vs_lambda}
shows the dependence of
$\rho\slash\alpha$ on $\lambda$ for $N=200$ and $a=5$ by circle $\circ$.
The results are averaged over 1000
samples of $\bm{y}$ and $\bm{A}$.
At a certain value of $\lambda$
that depends on $a$, AMP does not converge to fixed points,
for which the values of $\lambda$
are denoted by vertical black dashed lines in \Fref{fig:delta_vs_lambda}.
Understanding and
practically implementing AMP
requires stability analysis
of the fixed point.
We note that the dashed magenta lines
in \Fref{fig:delta_vs_lambda}
are the results obtained by the replica method that describes the asymptotic behaviour of AMP,
which is explained in Section \ref{sec:replica}.

\subsection{Macroscopic analysis: density evolution of message passing}
\label{sec:density_evolution}
Density evolution is a technique 
for analyzing the dynamical behaviour of BP through a macroscopic distribution of messages. We derive the density evolution for linear regression with the SCAD penalty in the case where the matrix $\bm{A}$ and vector $\bm{y}$ have random entries that are i.i.d., with mean $0$ and variances $1/M$ and $\sigma_{y}^2$, respectively.
The derivation is standard, and a similar approach is studied in inference setting where $\bm{y}$ is generated by linear transformation of a hidden true signal\cite{Krzakala2012, Barbier4}. Here we present the density evolution for the situation in which the distribution of $\bm{y}$ elements are not dependent on a true signal distribution but randomly and independently generated.

Combining equations \eref{eq:Ri}, \eref{A}, \eref{B}, \eref{g_out_SCAD}, \eref{g_out_p_SCAD}, and \eref{omega_df}, we obtain
the following expression of the quantity $R_{i}^{(t)}$:
\begin{equation}
R_{i}^{(t)}=\frac{1}{\sum_{\mu}A_{\mu i}^2}\sum_{\mu}\{A_{\mu i}y_{\mu}-A_{\mu i}\sum_{j\neq i}A_{\mu j}a_{j\rightarrow\mu}^{(t)}\},
\end{equation}
where the superscript $(t)$
denotes the
values at iteration step $t$.
For large dimensions $N\rightarrow\infty$, when looking at the macroscopic behaviour of messages, based on the TAP approximation \eref{TAP}, one could break the correlation between $a_{j\rightarrow\mu}^{(t)}$s for different $j$ and $\mu$. Then heuristically, one could see the variables $\{R_{i}^{(t)}\}$ as random variables with respect to the distribution of the basis matrix elements $A_{\mu i}$ and the data elements $y_{\mu}$;
hence, they can be effectively regarded as Gaussian random variables from the central limit theorem. 
At the current stage, the density evolution for our problem setting is heuristics. The rigorousness about density evolution of AMP is discussed in \cite{Bayati-Montanari2011,Bayati-Montanari2012}, and is beyond the scope of this paper.
Based on this approximation,
the mean of $R_i^{(t)}$
is given by
\begin{eqnarray}
\nonumber
\overline{R_{i}^{(t)}}&=&
\overline{\frac{1}{\sum_\mu A_{\mu i}^2}\left\{\sum_{\mu}A_{\mu i}y_{\mu}-\sum_{\mu}\sum_{j\neq i}A_{\mu i}A_{\mu j}a_{j\rightarrow\mu}^{(t)}\right\}}\\
&\simeq&\sum_{\mu}\overline{A_{\mu i}y_{\mu}}-\sum_{\mu}\sum_{j\neq i}\overline{A_{\mu i}A_{\mu j}a_{j\rightarrow\mu}^{(t)}}=0,
\label{eq:R_ave}
\end{eqnarray}
where $\overline{\cdots}$
denotes the average with respect to $\bm{A}$ and $\bm{y}$.
The variance of $R_i^{(t)}$,
defined as $E^{(t)}$,
is given by
\begin{eqnarray}
\nonumber
\overline{(R_{i}^{(t)})^2}&\simeq&\overline{\left\{\sum_{\mu}A_{\mu i}y_{\mu}-\sum_{\mu}\sum_{j\neq i}A_{\mu i} A_{\mu j}a_{j\rightarrow\mu}^{(t)}\right\}^2}\\
&\simeq&\sigma_{y}^2+\frac{1}{M}\sum_{j=1}^{N}\overline{(a_{j}^{(t)})^2}\equiv E^{(t)}\label{R_var}.
\end{eqnarray}
In the second step of approximation in \eref{R_var}, we ignore the terms of $O(1/\sqrt{N})$ in \eref{TAP}, for cases where $N, M\rightarrow\infty$. Based on this approximation, we replace $(a_{j\rightarrow\mu}^{(t)})^2$ with $(a_{j}^{(t)})^2$ and we can replace $\{A_{\mu j}^2\}$ with its expectation $M^{-1}$ from the law of large numbers. Based on the statistical property of $R_i^{(t)}$,
the marginal distribution at iteration $t+1$ is denoted by $m_{i}^{(t+1)}(x_i)$,
which is distributed as
\begin{eqnarray}
m_{i}^{(t+1)}(x_i)
&\simeq\frac{1}{\hat{Z}_i^{(t+1)}}P_{r}(x_i)\exp\left\{-\beta\frac{(x_i-z\sqrt{E^{(t)}})^2}{2(1+V^{(t)})}\right\},
\label{eq:dist_m}
\end{eqnarray}
where $z$ is a random Gaussian variable with zero mean and unit variance, and $\hat{Z}_i^{(t+1)}$ is the a normalization constant
of the marginal distribution
at step $t+1$.

\Eref{eq:dist_m} implies that
the marginal distribution
at step $t$
can be simply expressed in terms of two parameters
$V^{(t)}$ and $E^{(t)}$
at sufficiently large $N$.
Finally, by referring to (\ref{fa_general}), (\ref{fc_general}), and the definition of $E$ (\ref{R_var}), the density evolution equations are derived as
\begin{eqnarray}
V^{(t+1)}=\frac{1}{\alpha}\int Dz f_c(1+V^{(t)},z\sqrt{E^{(t)}}),\label{V_DE}\\
E^{(t+1)}=\frac{1}{\alpha}\int Dz f_a^{2}(1+V^{(t)},z\sqrt{E^{(t)}})+\sigma_{y}^2\label{E_DE},
\end{eqnarray}
where $\int Dz=\int_{-\infty}^\infty \frac{dz}{\sqrt{2\pi}}e^{-z^2\slash 2}$.
They describe how macroscopic parameters $E$ and $V$ evolve during the iterations of the BP algorithm. The density evolution equations are the same for the message passing and for the TAP equations, as the factors of ${{O}}(1/N)$ are ignored in the density evolution. The correspondence between density evolution and the replica symmetric solution is provided in Section \ref{sec:replica}.

In the current problem setting,
the fixed point of the density evolution equation is unique
within the range we observed
for any system parameters $\alpha$, $a$, and $\lambda$.

\subsection{Microscopic stability of AMP}
\label{sec: microscopic stability}

Even when the density evolutions
\eref{V_DE} and \eref{E_DE}
converge to a stationary state,
the convergence of the microscopic variables updated in AMP, such as $R_{i\rightarrow\mu}^{(t)}$, is not guaranteed.
We perform linear stability analysis of AMP's fixed points
to examine their local stability, which is a necessary condition for the global stability.
A similar analysis has been discussed in \cite{CDMA_Kabashima} for binary signals. Here, we provide a more general analysis.

Assuming the update of $R_{i\rightarrow\mu}^{(t)}$ is linearized around a fixed-point solution $R_{i\rightarrow\mu}$, using equations \eref{df:R_imu} and (\ref{TAP}) yields
\begin{eqnarray}
R_{i\rightarrow\mu}^{(t+1)} \nonumber
&=\frac{1}{\sum_{\gamma\neq\mu}{\cal{A}}_{\gamma\rightarrow i}^t}\left\{\sum_{\gamma\neq\mu}\frac{A_{\gamma i}y_{\mu}-A_{\gamma i}\sum_{j\neq i}A_{\gamma j}a_j}{V_{\mu}+1} \right\}\\
&\hspace{1.0cm}+\sum_{\gamma \neq\mu}\frac{A_{\gamma i}\sum_{j\neq i}A_{\gamma j}\frac{\partial f_{a}(\Sigma_{i}^2, R_{i})}{\partial R}}{V_{\mu}+1}R_{j\rightarrow\gamma}^{(t)}.
\end{eqnarray}
Therefore, the deviation from the fixed point can be expressed as
\begin{equation}
\delta R_{i\rightarrow\mu}^{(t+1)}=\frac{1}{V+1}\sum_{\gamma\neq\mu}A_{\gamma i}\sum_{j\neq i}A_{\gamma j}\frac{\partial f_{a}(\Sigma_{i}^2, R_{i})}{\partial R}\delta R_{j\rightarrow\gamma}^{(t)},
\label{delt_Rt}
\end{equation}
where we have replaced $V_{\mu}$ with the average $V$ and assumed that $M$ is large.
Assuming that $\delta R_{i\rightarrow\mu}^{(t)}$ are uncorrelated,
the summation including the basis matrix elements $\{A_{\mu i}\}$, which are i.i.d. random numbers, on the right-hand side of \eref{delt_Rt} leads to a Gaussian random number because of the central limit theorem. 
Furthermore, the correlations of $R_{i\rightarrow\mu}^{(t+1)}$ with respect to indices $\mu, i$, and $t$ are negligible,
because the right-hand side of (\ref{delt_Rt}) does not include the indices $\mu$ and $i$. These facts make it possible to analyse the stability of the fixed point
by observing the growth of the first and second moments of $\delta R_{i\rightarrow\mu}^{(t)}$ through each update.

We assume the self-averaging property for the derivation of the local stability condition,
where the quantity with respect to $\bm{y}$ and $\bm{A}$ converges to its expectation denoted by $\overline{\cdots}$.
By introducing the self-averaging property,
the macroscopic density evolution equation is connected with the local stability of AMP.
The first moment of the deviation from the fixed point $\delta R_{i\to\mu}^{(t)}$ is negligible at sufficiently large $M$ and $N$, since the average of $A_{\mu i}$ is zero.
The second moment of the deviation from the fixed point can be expressed as the square of $\delta R_{i\to\mu}^{(t)}$. The update relation is given by
\begin{eqnarray}
(\delta R_{i\rightarrow\mu}^{(t+1)})^2 &\simeq & \frac{1}{(V+1)^2}\overline{\left(\sum_{\gamma\neq\mu}A_{\gamma i}\sum_{j\neq i}A_{\gamma j}\frac{\partial f_{a}(\Sigma_{i}^2, R_{i})}{\partial R} \delta R_{j\rightarrow\gamma}^{(t)}\right)^2}\\\nonumber
&=& \frac{1}{\alpha(V+1)^2}\frac{1}{M}\sum_{\gamma\neq\mu}\left(\frac{1}{N}\sum_{j\neq i}\overline{(\frac{\partial f_{a}(\Sigma_{i}^2, R_{i})}{\partial R})^2}(\delta R_{j\rightarrow\gamma}^{(t)})^2\right)\\\nonumber
&\simeq& \frac{1}{\alpha(V+1)^2}\frac{1}{M}\sum_{\gamma\neq\mu}\left(\frac{1}{N}\sum_{j\neq i}\overline{(\frac{\partial f_{a}(\Sigma_{i}^2, R_{i})}{\partial R})^2}\right)\left(\frac{1}{N}\sum_{j\neq i}{(\delta R_{j\rightarrow\gamma}^{(t)})^2}\right), \label{delt_Rt_var}
\end{eqnarray}
where we have replaced the sample average of products with the product of the sample average in the last step, and this is valid when $N$ is large. In addition, the macroscopic variable $\overline{(\frac{\partial f_{a}(\Sigma_{i}^2, R_{i})}{\partial R})^2}$ can be expressed as
\begin{equation}
\overline{\left(\frac{\partial f_{a}(\Sigma_{i}^2, R_{i})}{\partial R}\right)^2}\simeq \int \textrm{D}z\left(\frac{\partial f_{a}(V+1, z\sqrt{E})}{\partial  (z\sqrt{E})}\right)^2
\end{equation}
regardless of $i$, where $V$ and $E$ are fixed-point values of $V^t$ and $E^t$ in the density evolution, respectively. For sufficiently large $M$ and $N$, $N^{-1}\sum_{j\neq i}(\delta R_{j\rightarrow\gamma}^{(t)})^2$ coincides with $(\delta R_{i\to\mu}^{(t)})^2$ itself,
because their dependency on $\mu$ can be ignored. Therefore, the squared deviation
is enlarged through the belief update, and the fixed-point solution becomes locally unstable if
\begin{equation}
\frac{1}{\alpha (V+1)^2}\int \textrm{D}z\left(\frac{\partial f_{a}(V+1, z\sqrt{E})}{\partial (z\sqrt{E})}\right)^2>1.
\label{stability_BP}
\end{equation}
Using \eref{stability_BP}, one can specify the parameter region where AMP's fixed point is locally stable for any regularization by inserting the specific function form of $f_a$ for a given regularization. We remind the reader that the convergent $f_a$ is the estimated value for $x$. 
In general, the analytic expression of $f_a$ is not derived for arbitrary regularizations, but the computation of $f_a$ is possible practically by numerically solving \eref{eq:f_a_beta_infty}. One of the nonconvex regularizations except SCAD that provide the analytical form of $f_a$ is MCP; hence, we consider that the analysis shown here is directly applied to MCP.
We will show the correspondence of the AMP's local stability condition and the de Almeida--Thouless condition derived from the replica method in Section \ref{sec:replica}.

To improve the convergence trajectory, employing an appropriate damping factor for each $a_i$ and $v_i$ in each update is a standard procedure. Note that the stability analysis is for the fixed point of the algorithm, meaning that once the algorithm reaches the fixed point, it becomes stable. The damping factor helps to stabilize the algorithm's trajectory to reach the fixed point by shortening the changing distance of the vectors $\bm{a}$ and $\bm{v}$ in each step. 

\section{Replica analysis}
\label{sec:replica}
The replica method provides the performance of the optimal algorithm for problem \eref{eq:SCAD_LR}. The asymptotic property of AMP's fixed point presented in the previous section can also be analytically derived independently using the replica method.
The replica method is a heuristics, 
but its validness is shown through its application to problems in information science \cite{Tanaka,CS_KWT,RFG}.
Further, its rigorousness is proved in several problems \cite{Barbier,RS_exact}.
The replica method provides the physical meanings of the density evolution and the stability of AMP.
The basis for the analysis is the
free energy density defined by
\begin{eqnarray}
f=-\lim_{\beta\to\infty}\frac{1}{M\beta}\overline{\ln Z_{\beta}(\bm{y},\bm{A})},
\end{eqnarray}
which corresponds to the expectation of the
minimum value of the energy $e(\bm{x}|\bm{y},\bm{A})$.

The expectation according to \eref{eq:P_yA} is implemented by the replica method
based on the following identity:
\begin{eqnarray}
\overline{\ln Z_{\beta}(\bm{y},\bm{A})}=
\lim_{n\to 0}\frac{\overline{Z_{\beta}^n(\bm{y},\bm{A})}-1}{n},
\label{eq:replica}
\end{eqnarray}
which is employed for the analysis of sparse  regularizations
in various problems \cite{Kabashima2009,Sakata2016,Xu2016,Xu2013}.
We briefly summarise the replica analysis for the SCAD penalty.
The detailed explanation is
given in \ref{sec:replica_SCAD}.

\begin{figure}
  \centering
    \includegraphics[width=3in]{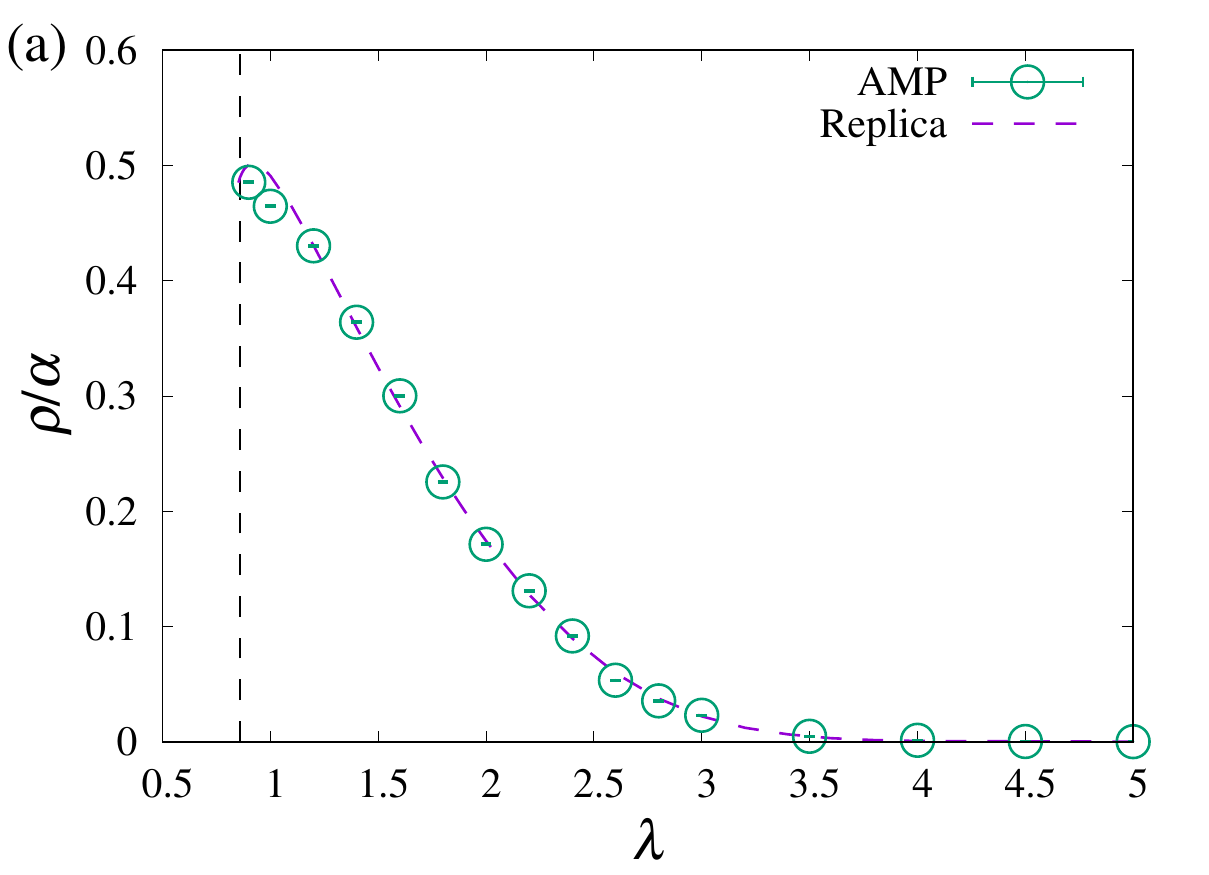}
    \includegraphics[width=3in]{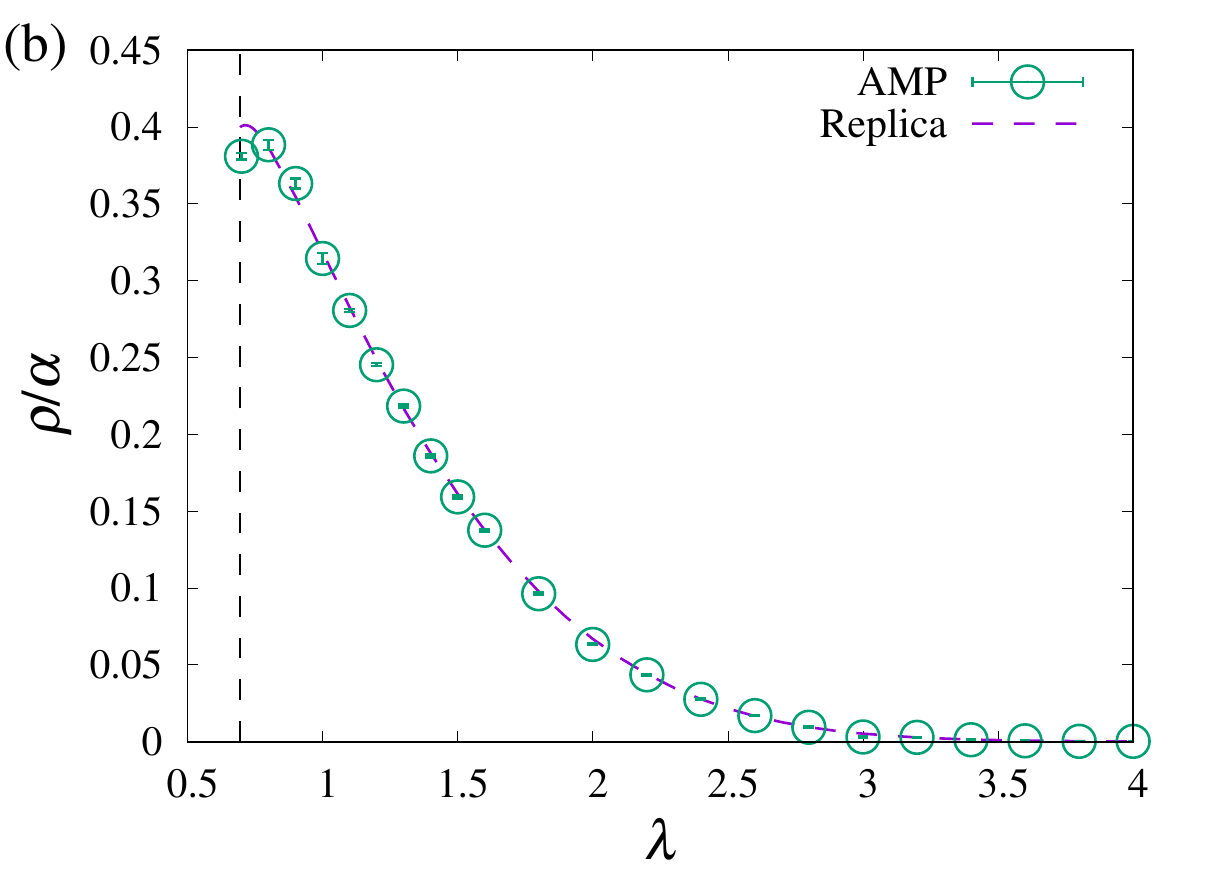}
    \caption{Dependence of $\rho\slash\alpha$ on $\lambda$ at SCAD-AMP's fixed point for $a=5$ with (a) $\alpha=0.1$ and (b) $\alpha=0.5$.
Circles are obtained by the SCAD-AMP algorithm. The dashed magenta lines represent
the result derived by the replica method, and
SCAD-AMP does not converge in the left-hand side region of the black dashed lines.}
\label{fig:delta_vs_lambda}
\end{figure}

We focus on the $N\to\infty$ and $M\to\infty$ limits
by keeping $M\slash N=\alpha\sim O(1)$.
Under the replica symmetric (RS) assumption,
the free energy density is given by
\begin{eqnarray}
f=\mathop{\rm extr}_{Q,\chi,\hat{Q},\hat{\chi}}\Big\{\frac{\alpha(Q+\sigma_y^2)}{2(1+\chi)}-\frac{\alpha(Q\hat{Q}-\chi\hat{\chi})}{2}
+\frac{1}{2}\xi(\hat{Q},\hat{\chi})\Big\},
\label{eq:free_energy}
\end{eqnarray}
where ${\rm extr}_{Q,\chi,\hat{Q},\hat{\chi}}$
denotes extremization with respect to the
variables $\{Q,\chi,\hat{Q},and \hat{\chi}\}$,
which are given by
\begin{eqnarray}
Q&=\frac{1}{\alpha}\frac{\partial \xi}{\partial \hat{Q}}\\
\chi&=-\frac{1}{\alpha}\frac{\partial \xi}{\partial\hat{\chi}}\\
\hat{Q}&=\frac{1}{1+\chi}\label{eq:hat_Q}\\
\hat{\chi}&=\frac{Q+\sigma_y^2}{(1+\chi)^2}\label{eq:h_chi}
\end{eqnarray}
at the extremum.
The function $\xi(\hat{Q},\hat{\chi})$ is given by
\begin{eqnarray}
&\xi(\hat{Q},\hat{\chi})=2\int Dz f_\xi(\sqrt{\hat{\chi}}z,\hat{Q})\label{eq:f_pi_def}
\\
&f_\xi(\sqrt{\hat{\chi}}z,\hat{Q})=\min_x\Big(\frac{\hat{Q}}{2}x^2-\sqrt{\hat{\chi}}zx+J_{\lambda,a}(x)\Big),\label{eq:one_body}
\end{eqnarray}
where $\sqrt{\hat{\chi}}z$
is the random field that effectively represents the
randomness introduced by $\bm{y}$
and $\bm{A}$.
It is shown that
$Q$ and $\chi$ relate to the
physical quantities at the extremum as
\begin{eqnarray}
Q&=\lim_{M\to\infty}\frac{1}{M}\overline{\hat{\bm{x}}^{\rm T}(\bm{y},\bm{A})\hat{\bm{x}}(\bm{y},\bm{A})}\\
\chi&=\lim_{\beta\to\infty}\lim_{M\to\infty}\frac{\beta}{M}\sum_{i=1}^N\overline{\left(\langle x_i^2\rangle_\beta-\langle x_i\rangle_\beta^2\right)},\label{eq:chi_physical}
\end{eqnarray}
where the superscript ${\rm T}$ denotes the matrix transpose.
The solution of $x$ concerned with the effective single-body problem \eref{eq:one_body},
denoted by $x^*(z;\hat{Q},\hat{\chi})$
for SCAD regularization, is given by
\begin{eqnarray}
x^*(z;\hat{Q},\hat{\chi})=
\frac{\sqrt{\hat{\chi}}z-\lambda{\rm sign}(z)}{\hat{Q}}{\rm I}
+ \frac{\sqrt{\hat{\chi}}z-{\frac{a\lambda}{a-1}{\rm sign}(z)}}{\hat{Q}-(a-1)}{\rm II}+\frac{\sqrt{\hat{\chi}}z}{\hat{Q}}{\rm III},
\label{eq:single_body_SCAD}
\end{eqnarray}
where
\begin{eqnarray}
{\rm I}&=\Theta(|z|\geq\sqrt{2}\theta_1)\Theta(|z|< \sqrt{2}\theta_2)\\
{\rm I\!I}&=\Theta(|z|\geq \sqrt{2}\theta_2)\Theta(|z|<\sqrt{2}\theta_3)\\
{\rm I\!I\!I}&=\Theta(|z|\geq \sqrt{2}\theta_3),
\end{eqnarray}
and the thresholds are given by
$\theta_{1}=\lambda\slash\sqrt{2\hat{\chi}}$,
$\theta_{2}=\lambda(\hat{Q}+1)\slash\sqrt{2\hat{\chi}}$, and
$\theta_{3}=a\lambda\hat{Q}\slash\sqrt{2\hat{\chi}}$,
respectively.
The behaviour of $x^*$ is the same as that shown in \Fref{fig:SCAD_single_body}. Using $x^*$,
we can represent the quantities $Q$ and $\chi$
in the replica method as
\begin{eqnarray}
Q&=\frac{1}{\alpha}\int Dz (x^*)^2\\
\chi&=\frac{1}{\alpha}\int Dz\frac{\partial x^*}{\partial(\sqrt{\hat{\chi}}z)}.
\end{eqnarray}
The solution $x^*$ is statistically equivalent to the solution of the
original problem \eref{eq:SCAD_LR},
which takes a nonzero value when $|z|\geq\sqrt{2}\theta_1$, where $z\sim {\cal N}(0,1)$.
Therefore,
the density of the nonzero component is given by
the probability that $|z|\geq\sqrt{2}\theta_1$:
\begin{equation}
\rho={\rm erfc}(\theta_1),
\label{eq:rho}
\end{equation}
where ${\rm erfc}(\theta)=\frac{2}{\sqrt{\pi}}\int_\theta^\infty dz\exp(-z^2).$
The dependence of $\rho\slash\alpha$ on $\lambda$
derived by the replica method is shown in \Fref{fig:delta_vs_lambda} by dashed lines.
The results of the replica method match those of AMP for a sufficiently
large system size,
which is mathematically supported
by considering the correspondence between the variables used in AMP and the replica method as explained in \cite{Barbier2,Barbier3}.
In addition,
the RS saddle point equation has
a unique solution; the first-order transition does not occur. Therefore, the local stability of the AMP's fixed points indicates their global stability.

A comparison of $f_a$ in AMP \eref{eq:f_a_beta_infty} and $x^*$ in replica analysis \eref{eq:one_body}
shows that they are equivalent to each other
on the basis of the correspondences
\begin{eqnarray}
\Sigma^2&\leftrightarrow\hat{Q}^{-1},
\label{eq:Sigma2_AMP_to_replica}\\
R_i\slash\Sigma^2&\leftrightarrow\sqrt{\hat{\chi}}z.\label{eq:R_i_AMP_to_replica}
\end{eqnarray}
From equations (\ref{Sigma_SCAD}) and (\ref{eq:hat_Q}),
the correspondence
\eref{eq:Sigma2_AMP_to_replica}
is equivalent to that between $V$ in AMP and $\chi$ in replica analysis.
This is consistent with the definition of $V$ and the physical meaning of $\chi$ given by \eref{eq:chi_physical}.
\Eref{eq:R_i_AMP_to_replica}
implies that the distribution of
$R_i\slash\Sigma^2$
with respect to $\bm{y}$ and $\bm{A}$ is
represented by a Gaussian distribution with variance $\hat{\chi}$.
From the calculation explained in \ref{sec:replica_SCAD},
it is shown that $\hat{\chi}$
corresponds to the representation error defined by
\begin{eqnarray}
{\rm err}=\frac{1}{M}\overline{||\bm{y}-\bm{A}\hat{\bm{x}}(\bm{y},\bm{A})||_2^2},
\label{eq:def_err}
\end{eqnarray}
which evaluates the accuracy of the expression of data using the estimated sparse expression.
The variance of $R_i$
is defined as $E$;
hence,
\begin{eqnarray}
E\leftrightarrow\Sigma^2\hat{\chi}=(1+V)\hat{\chi}\leftrightarrow (Q+\sigma_y^2).
\end{eqnarray}
This correspondence implies that
density evolutions \eref{V_DE} and \eref{E_DE} can be regarded as recursively solving the saddle point equations
$\chi$ and $(Q+\sigma_y^2)$,
respectively, under the RS assumption.

\subsection{de Almeida--Thouless condition for the replica symmetric phase}
\label{sec:AT_RS}
The RS solution discussed so far loses local stability under perturbations that
break the symmetry between replicas in a certain parameter region. This phenomenon is known as de
Almeida--Thouless (AT) instability \cite{AT}, and it occurs when
\begin{eqnarray}
\frac{1}{\alpha(1+\chi)^2}\int Dz\left(\frac{\partial x^*}{\partial(\sqrt{\hat{\chi}}z)}\right)^2 >1
\label{AT_replica}
\end{eqnarray}
holds,
as explained in \ref{sec:replica_AT}.
By applying this to the
minimizer of the single-body problem \eref{eq:single_body_SCAD},
we get the AT instability condition
for the SCAD penalty as
\begin{eqnarray}
\frac{\rho}{\alpha}+\left\{\left(\frac{\hat{Q}}{\hat{Q}-\frac{1}{a-1}}\right)^
{2}-1\right\}\frac{\gamma}{\alpha}>1,
\label{eq:AT}
\end{eqnarray}
where $\gamma={\rm erfc}(\theta_2)-{\rm erfc}(\theta_3)$.
At $a\to\infty$, the AT instability condition for SCAD reduces to that
for $\ell_1$ regularization:
$\rho\slash\alpha>1$.
Here, $\rho\slash\alpha$ corresponds to the ratio of the number of variables to be estimated
to the number of known variables.
Hence, $\rho\slash\alpha\leq 1$ is the physically meaningful region,
and the Lasso is always stable against the symmetry breaking perturbation in the physical parameter region. Therefore, it is considered that the RS/RSB transition studied here is induced by finite $a$, which induces nonconvexity to the regularization.
The second term of \eref{eq:AT}
is the characteristic of SCAD
 regularization.
By definition, $\gamma\slash \alpha$ is the probabilistic weight of the transit region (region ${\rm I\!I}$ of \Fref{fig:SCAD_single_body}), and
$\hat{Q}\slash(\hat{Q}-(a-1)^{-1})$ denotes
the ratio between the variances
in the transient region and
those in other regions (equation (\ref{eq:single_body_SCAD})).
\Eref{eq:AT} implies that as
the transient region is extended
or as the variance of the estimates
in the transient region becomes smaller,
the RS solution
is likely to be unstable.

Based on the correspondence between AMP and the RS saddle point discussed so far,
the AT instability
condition \eref{AT_replica} is coincident with the
instability condition of density evolution given by \eref{stability_BP}, since $f_a$ in \eref{stability_BP} and $x^*$ in \eref{AT_replica} are the same and both mean the estimated value of $x$.
Similar correspondence has been shown in the BP algorithm for CDMA \cite{CDMA_Kabashima}.
We note that the correspondence
between AMP's local stability and the RS/RSB transition holds regardless of the type of  regularization, because regularization dependence only appears in $f_a$ in \eref{stability_BP} and $x^*$ in \eref{AT_replica} and they are the same. Hence,
the result is valid for other regularizations where the RS/RSB transition appears.

\section{Phase diagram and representation error}
\label{sec:phasetransition_error}
\begin{figure}
\centering
\includegraphics[width=3in]{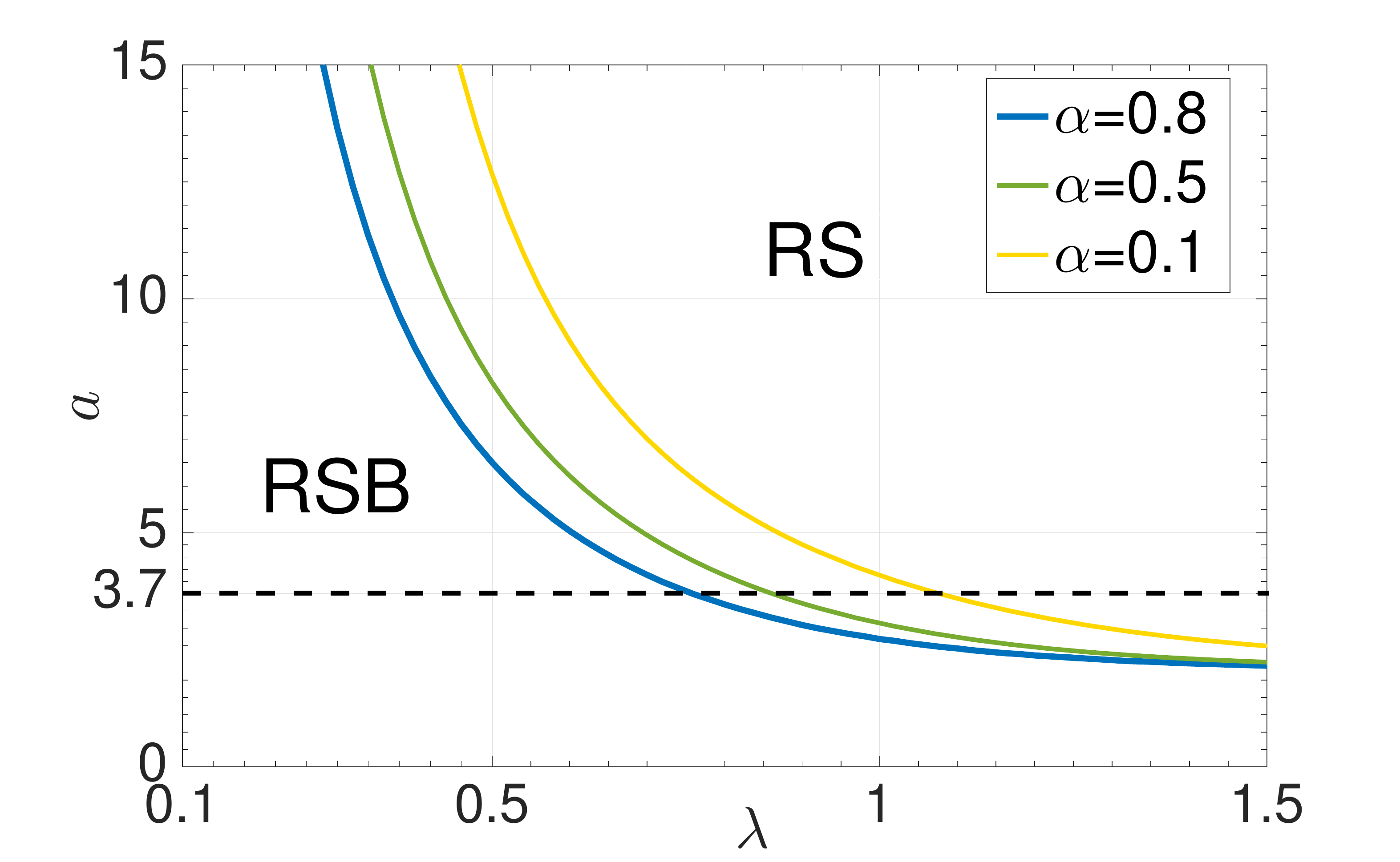}
\caption{Phase diagram of the $a-\lambda$ plane at $\alpha=0.8, \alpha=0.5$, and
 $\alpha=0.1$.
The conventional value $a=3.7$ is indicated by the horizontal dashed line.}
\label{fig:phase}
  \end{figure}

\Fref{fig:phase} shows the phase diagram
on the $\lambda-a$ plane
for (a) $\alpha=0.1$ and (b) $\alpha=0.5$.
The range of parameters that gives the RS phase shrinks
as $\alpha$ decreases.
The conventional value $a=3.7$ \cite{SCAD},
which is indicated by horizontal lines
in \Fref{fig:phase},
is considered as an appropriate value of $a$
to obtain RS stability at a sufficiently large $\lambda$.
As $\lambda$ decreases,
the value of $a$ for achieving the RS phase diverges,
consistent with the shape of SCAD regularization.

\begin{figure}
\centering
\begin{minipage}{0.48\textwidth}
  \includegraphics[width=3in]{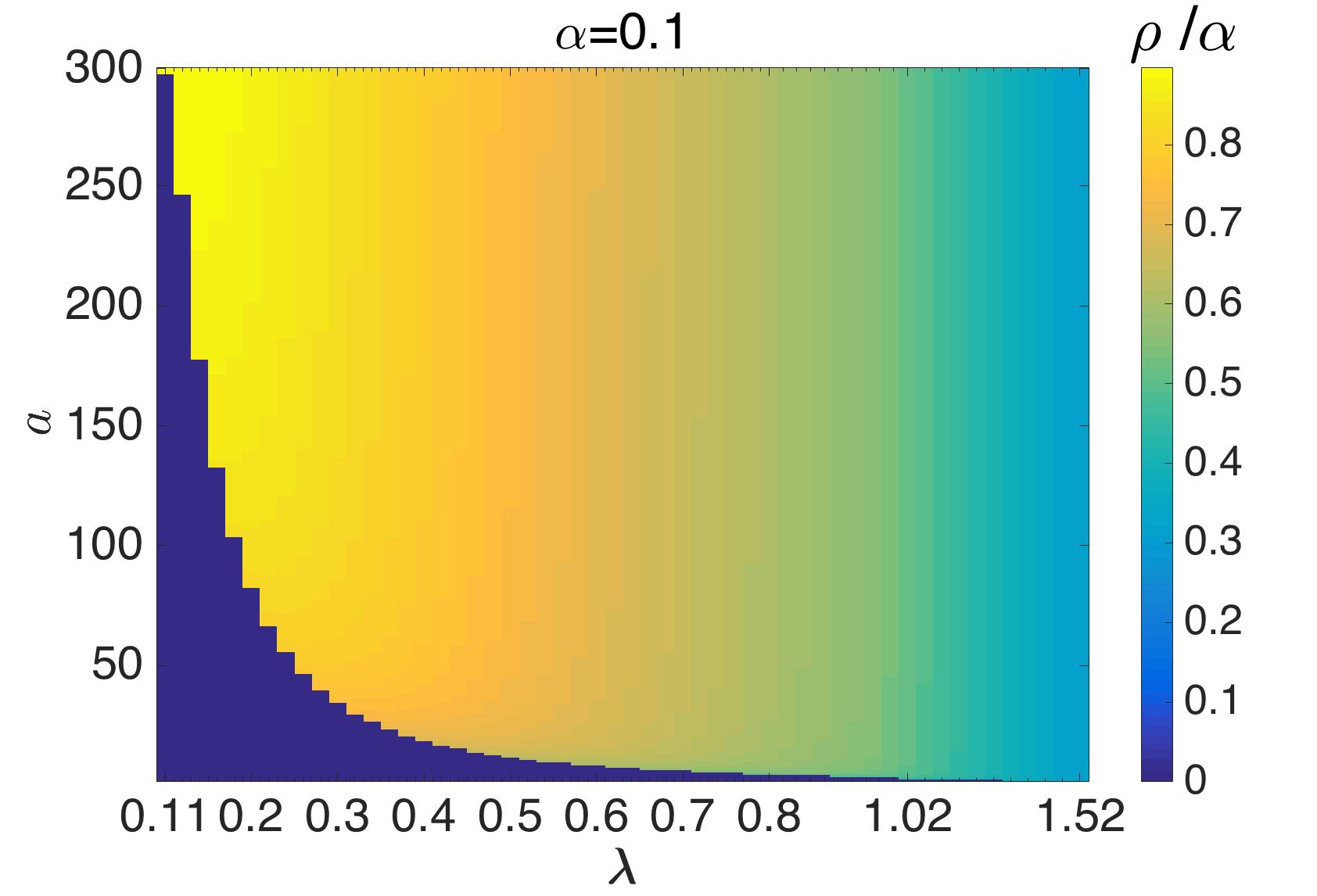}
  \centering (a)
  \end{minipage}
  \begin{minipage}{0.48\textwidth}
  \includegraphics[width=3in]{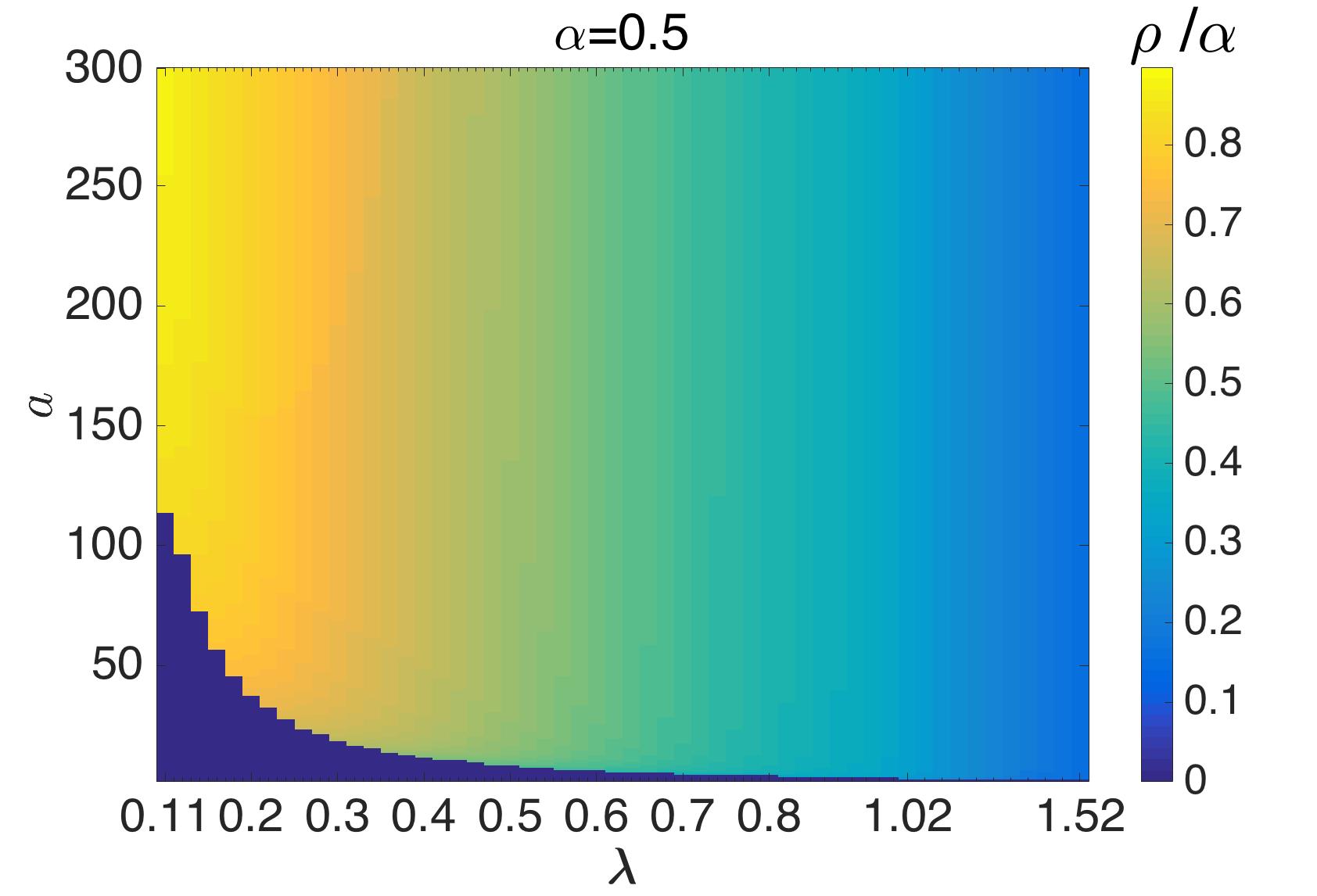}
  \centering (b)
  \end{minipage}
  \caption{Parameter dependence of the sparsity at (a) $\alpha=0.1$ and (b) $\alpha=0.5$ in the RS phase. The figures were created by presenting the elements of matrix $\rho(a,\lambda)\slash\alpha$ in a colour map, resulting in block-like shapes.}
\label{fig:sparsity}
\end{figure}

\Fref{fig:sparsity} shows the parameter dependence of the sparsity $\rho\slash\alpha$
in the RS phase.
The sparsity is nearly controlled by $\lambda$;
as $\lambda$ increases, the nonzero components in the estimate are enhanced.
The sparsity slightly depends on $a$ compared with $\lambda$,
but as $a$ decreases, the number of nonzero components increases.
This result is consistent with the form of the SCAD regularization \eref{eq:SCAD_def},
because the gradient around the origin in region $|x|\leq\lambda$, which is a key aspect of the sparsity, 
is governed by $\lambda$.

As the number of nonzero components decreases, the
representation performance of the data is generally degraded.
There is a trade-off relationship between the simplicity and
the accuracy of the expression.
We quantify this relationship
by regarding the representation error \eref{eq:def_err}
as a function of sparsity $\rho\slash\alpha$. 
As mentioned in Section \ref{sec:replica} and
explained in \ref{sec:replica_SCAD},
the representation error is given by $\hat{\chi}$ in the replica method.
Hence, we obtain the dependency of the representation error on the sparsity by solving \eref{eq:h_chi} and \eref{eq:rho}.

\begin{figure}
  \centering
    \includegraphics[width=3in]{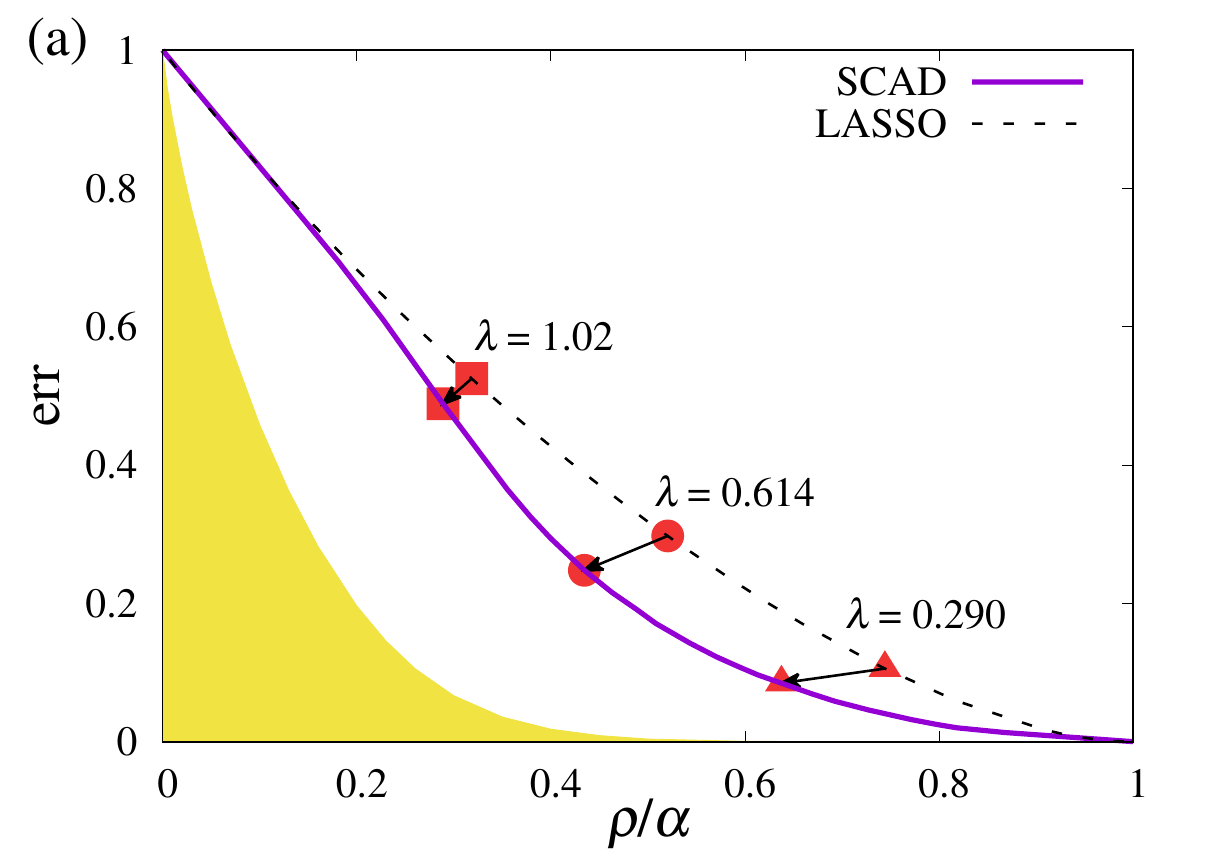}
    \includegraphics[width=3in]{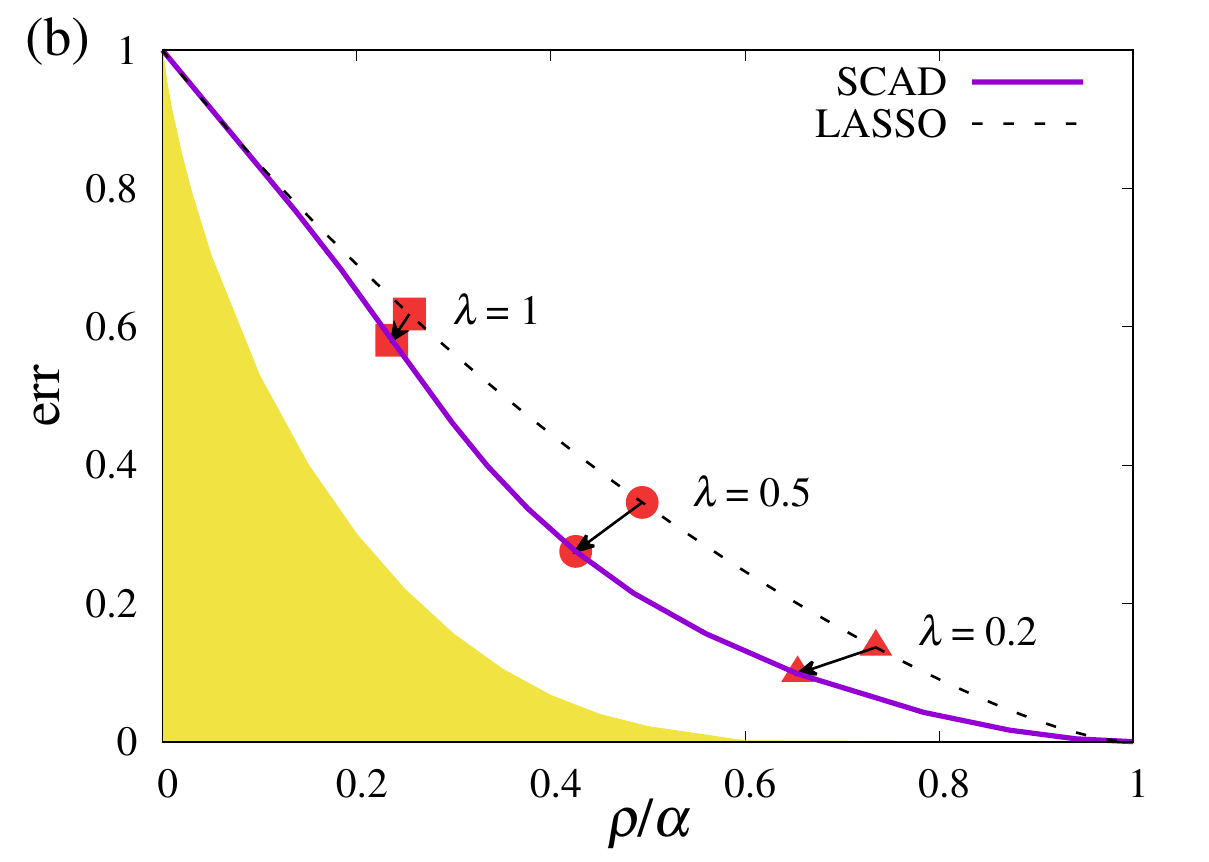}
    \caption{Relationship between the representation error and the normalized sparsity $\rho/\alpha$ for (a) $\alpha=0.5$ and (b) $\alpha=0.8$.
The LASSO corresponds to the limit $a\to\infty$,
and the shift of the error curve by decreasing $a$ is indicated by arrows.
The shaded regions are regions that are not achievable for any compression method \cite{Nakanishi-Ohno2016}.}
\label{fig:rate_distortion}
\end{figure}

\Fref{fig:rate_distortion} shows
the representation error as a function of the sparsity at $\alpha=0.5$ and $\alpha=0.8$.
For comparison,
the results obtained by $\ell_1$ regularization,
which corresponds to $a\to\infty$,
are shown by dashed lines,
and the regions that are not achievable
for any estimation method \cite{Nakanishi-Ohno2016}
are shaded.
At each value of $\lambda$,
we find the value of $a$ that gives the smallest representation
error.
The shift of the representation error curve associated with the
decrease in $a$
is shown by arrows in \Fref{fig:rate_distortion},
where the minimum values of
$a$ are (a) $a=3$, 6, and 20 for $\lambda=0.290$, 0.614, and 1.02, respectively,
and (b) $a=2.739$, 6.51, and 25
for $\lambda=0.2$, 0.5, and 1,
respectively.
The representation error monotonically decreases as $a$ decreases. Hence,
when we restrict the SCAD parameters to be within the replica symmetric region,
the smallest representation error is obtained at the RS/RSB boundary.
In addition, the sparsity slightly decreases as $a$ decreases.
Hence, the sparsest and
most accurate expression is obtained by
decreasing $a$ to be on
the RS/RSB boundary.

\section{Correspondence between RS/RSB transition and
convergence of coordinate descent:
a conjecture}
\label{sec:Coordinate_Descent}

For the SCAD penalty,
it is shown that
the CD algorithm
is valid in a certain parameter region \cite{Breheny2011}.
CD is the component-wise minimization of energy density \eref{eq:energy}
while all other components
are fixed; it cycles through all
parameters until convergence is reached.
Let us denote the
estimate at step $t$
of CD as $\hat{\bm{x}}^{(t)}$
and define the residue at step 0
as $\bm{r}^{(0)}=\bm{y}-\bm{A}\hat{\bm{x}}^{(0)}$.
CD for SCAD is given by
the cyclic update of
one component of the
estimate $\hat{\bm{x}}$
according to
the following equations \cite{Breheny2011}:
\begin{eqnarray}
z_j^{(t+1)} &= \bm{A}_j^{\rm T}\bm{r}^{(t)}+\hat{x}_j^{(t)}\\
\hat{x}_j^{(t+1)}&=\left\{\begin{array}{ll}
S(z_j^{(t+1)},\lambda) & {\rm for}~|z_j^{(t+1)}|\leq 2\lambda\\
\displaystyle\frac{S(z_j^{(t+1)},a\lambda\slash(a-1))}{1-(a-1)^{-1}} & {\rm for}~2\lambda<|z_j^{(t+1)}|\leq a\lambda \\
z_j^{(t+1)} & {\rm for}~|z_j^{(t+1)}|>a\lambda \\
\end{array}\right.
\\
\bm{r}^{(t+1)}&=\bm{r}^{(t)}-(\hat{x}_j^{(t+1)}-\hat{x}_j^{(t)})\bm{A}_j,
\end{eqnarray}
where $S$ is the soft-thresholding function
defined by
\begin{eqnarray}
S(z,\lambda)=\left\{\begin{array}{ll}
z-{\rm sign}(z)\lambda & {\rm for}~ |z|>\lambda \\
0     &{\rm otherwise}
\end{array}\right.,
\end{eqnarray}
and $\bm{A}_j$ denotes the $j$-th column vector of $\bm{A}$.
A sufficient condition for the
convergence of CD to the globally stable state is
proposed as \cite{Breheny2011}
\begin{eqnarray}
a>1+{c(\lambda,a;\bm{A})}^{-1}.
\label{eq:sufficient}
\end{eqnarray}
Here, $c(\lambda,a;\bm{A})$ is the minimum eigenvalue of
the Gram matrix $\bm{A}_{K_{\lambda-d\lambda,a}}^{\rm T}\bm{A}_{K_{\lambda-d\lambda,a}}$,
where $\bm{A}_{K}$ denotes the submatrix of $\bm{A}$ that consists of columns in the set $K$, and 
$K_{\lambda,a}$ is the support at $\lambda$ and $a$.
The infinitesimal quantity $d\lambda(>0)$ is set such that
$|K_{\lambda-d\lambda,a}|-|K_{\lambda,a}|=1$.
However, the necessary and sufficient condition for
the convergence of CD is
not known.
Here, we suggest that
the convergence conditions of
CD and AMP are equivalent
from numerical observation.

\begin{figure}[t]
\centering
\includegraphics[width=3in]{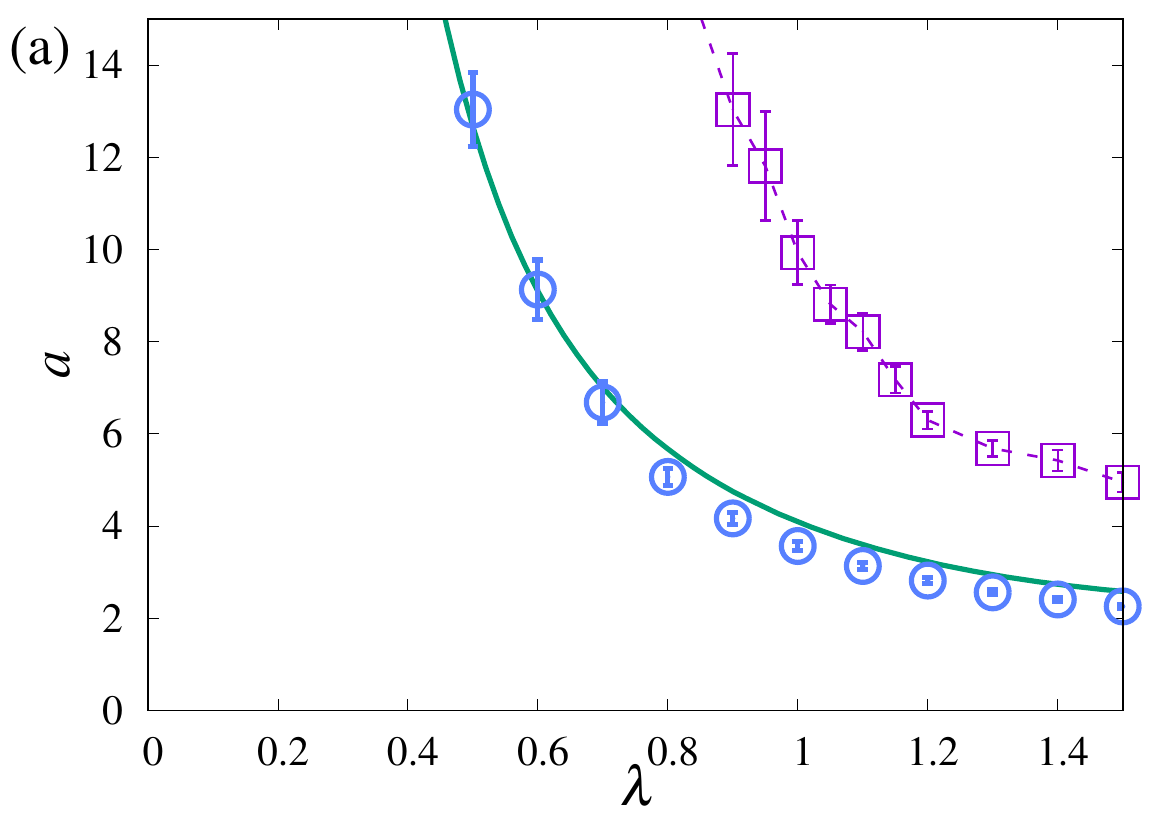}
\includegraphics[width=3in]{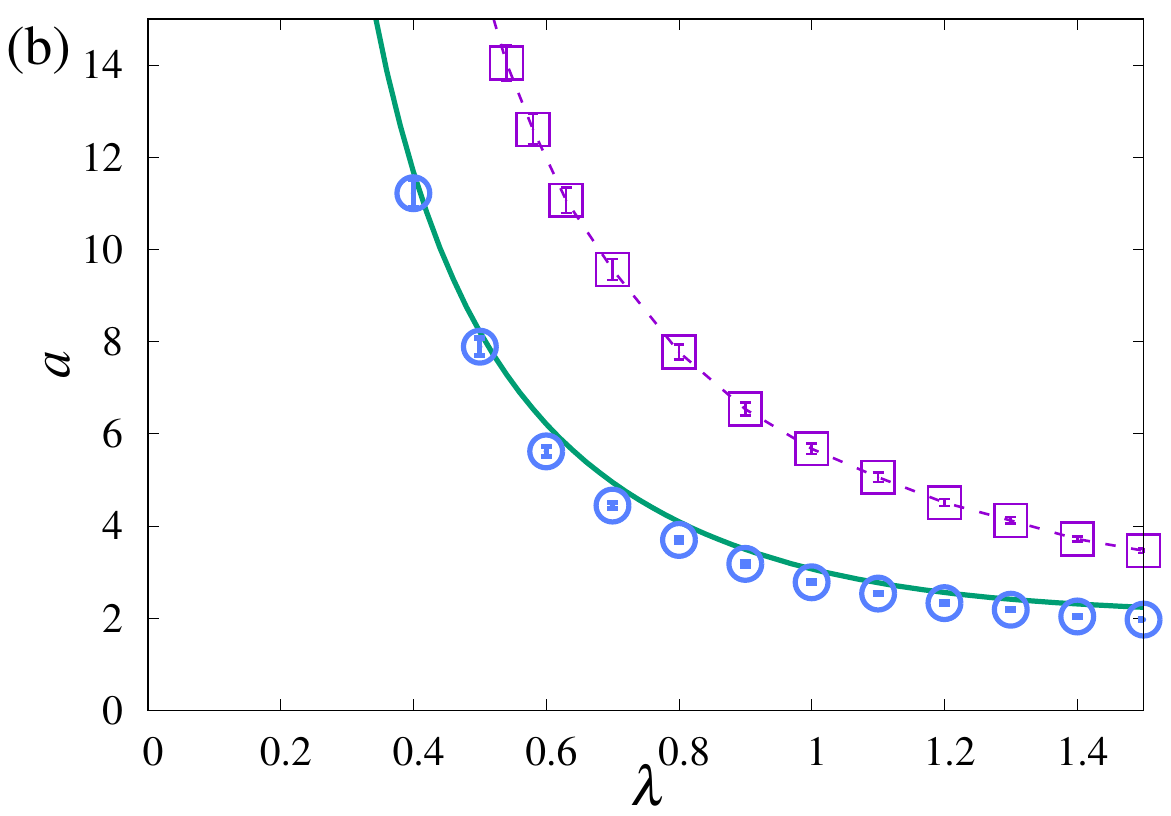}
\caption{Comparison between
RS/RSB phase boundary (solid lines) and
typically solvable limit of CD
(denoted by $\circ$)
for (a) $N=200,~M=20~(\alpha=0.1)$ and (b) $N=200,~M=100~(\alpha=0.5)$.
The sufficient condition for the convergence of CD is
shown by solid lines with $\square$.}
\label{fig:CD}
\end{figure}

To check the convergence of CD, we prepare $m$ random initial conditions.
The fixed point of CD starts from the $k$-th initial condition
under the fixed data $\bm{y}$ and the predictor matrix $\bm{A}$
is denoted as $\hat{\bm{x}}_k^{\rm CD}(\bm{y},\bm{A})$.
We compute the
average of the
differences between the fixed points as
\begin{eqnarray}
d(\bm{y},\bm{A})=\frac{1}{\frac{m(m-1)}{2}}\sum_{k<l}||\hat{\bm{x}}^{\rm CD}_k(\bm{y},\bm{A})-\hat{\bm{x}}^{\rm CD}_l(\bm{y},\bm{A})||^2_2.
\end{eqnarray}
The uniqueness of the stable solution for CD is
indicated by $d(\bm{y},\bm{A})=0$.
We define $a^*(\bm{y},\bm{A};\lambda)$
as the minimum value of $a$
that satisfies $d(\bm{y},\bm{A})=0$ for each $\lambda$.
In other words,
at $a<a^*(\bm{y},\bm{A};\lambda)$,
CD does not always attain the global minimum.
In \Fref{fig:CD},
the averaged value of $a^*(\bm{y},\bm{A};\lambda)$,
which indicates a typical solvable
limit,
over 100 samples of i.i.d. Gaussian random variables $\bm{y}$
and $\bm{A}$ are shown
by circles $\circ$
for (a) $N=200,~M=20~(\alpha=0.1)$ and (b) $N=200$, $M=50~(\alpha=0.5)$,
where we set $m=100$.
For comparison,
we show the convergence condition
$\eref{eq:sufficient}$
averaged over the same $\bm{y}$
and $\bm{A}$
by squares $\square$,
and the RS/RSB boundary \eref{eq:AT}.
As shown in \Fref{fig:CD},
the RS/RSB transition is an approval indicator of the convergence of CD,
although complete correspondence between them has not been mathematically proved.
From the physical consideration,
the RS/RSB transition is supposed to
correspond to the appearance of the
local minima, whose number is
an exponential order of the system size.
This view of the RS/RSB transition
is consistent with the result shown in \Fref{fig:CD}.

\section{Conclusion and discussion}
\label{sec:conclusion}

We studied a linear regression problem
under SCAD regularization.
We derived AMP for SCAD regularization and identified the stability condition
of AMP's fixed points.
This stability condition corresponds to the AT condition
derived by the replica method,
and the correspondence between the stability of AMP and
that of the RS solution was indicated.
The stability analysis does not depend on the form of the regularization;
hence, the correspondence holds for other
regularizations that exhibit the RS/RSB transition.
In addition, we identified the replica symmetric phase on the $\lambda-a$
parameter plane,
and quantified the relationship between the representation error
and the sparsity.
Furthermore, we theoretically showed that, for 
data $\bm{y}$ and the basis matrix $\bm{A}$ consisting of i.i.d. Gaussian random variables,
SCAD regularization typically provides a more accurate and sparser expression compared with $\ell_1$ regularization in the RS phase. In addition, the smallest representation error is obtained at the RS/RSB boundary for each $\lambda$.

Our result not only shows that the nonconvexity of the sparse  regularization
does not always result in the instability of the replica symmetry
but also highlights the potential of message passing algorithms for nonconvex regularizations.
Because the unclear condition of global stability is the main concern for nonconvex sparse regularization in practice, identifying and using the RS region of other nonconvex sparse regularizations is a topic of interest for future work.
Further, AMP for nonconvex regularizations is applicable to the inference setting, in which
$\bm{y}$ is generated as $\bm{y}=\bm{A}\bm{x}^0+\bm{\epsilon}$ using true signal $\bm{x}^0$ and noise $\bm{\epsilon}$. The reconstruction performance of $\bm{x}^0$ should be studied for understanding the contribution of the nonconvexity. For such cases, one should note that SCAD is used as a regularization and not a true prior distribution of the hidden true signal. Therefore, deciding the optimal parameters in SCAD for learning the signal from data $\bm{y}$ is not a trivial problem, since the true distribution of the signal $\bm{x}^0$ is unknown. 

Determination of the regularization parameters, which is also called model selection, while the true generating model of data is unknown, is one of the important problems of penalized regression in statistics.
In the case of the Lasso, model selection is implemented by using the Akaike's information criterion (AIC), which corresponds to an unbiased estimator of the prediction error \cite{GDF}. However, it is not straightforwardly applied to other regularizations. In fact, it is known that AIC does not correspond to unbiased estimators of prediction error for other regularizations \cite{Sakata2016}.
The development of the model selection method in conjunction with AMP is required for the practical usage of the nonconvex penalties.

In this work,
we focused on the case in which data $\bm{y}$ and predictor matrix $\bm{A}$ are i.i.d. Gaussian random variables.
Our discussion is applicable to non-Gaussian $\bm{A}$ and $\bm{y}$ that satisfy the following conditions: the correlation between components is negligible, each component of $\bm{A}$ has zero mean and variance $1\slash M$, and each component of $\bm{y}$ has zero mean and variance $\sigma^2_{y}$. However, in dealing with general data, a nonzero mean might be considered. As the same for other standard AMP algorithms, our algorithm is applicable to such cases by introducing a simple modification
that leads to a zero-mean matrix and vector as $\bm{y}-\bar{\bm{y}}\sim(\bm{A}-\bar{\bm{A}})\bm{x}$, where $\bar{\bm{y}}\equiv (\frac{1}{M}\sum_{\mu}y_{\mu})\bm{1}$, $\bm{1}$ denotes the $M$-dimensional identity vector, and $\bar{\bm{A}}$ is the $M\times N$ matrix where the $i$-th column is given by the values $\bar{A}_{i}\equiv (1/M)\sum_{\mu}A_{\mu i}$.
For a more general case where
the correlation between the components of $\bm{A}$ and $\bm{y}$ is not negligible,
the convergence of AMP and the validity of the approximations introduced into AMP are interesting problems to be resolved. The incorporation of nonconvex regularization into variants of AMP \cite{Ma,VAMP} where the convergence is improved merits further discussion.

\ack
The authors would like to thank Yoshiyuki Kabashima
for his insightful discussions and comments. This work is supported by JSPS KAKENHI No.16K16131,
Shiseido Female Researcher Science Grant (AS), and the Academy of Finland through its Center of Excellence in Computational Inference Research (COIN)(YX).

\newpage
\appendix
\section{Derivation of SCAD-AMP algorithm}
\label{app:derivation_of_SCAD_AMP}

We derive the AMP algorithm from the general form \cite{Rangan2011, Kabashima2004GAMP, Xu2014} to the specific form for the SCAD penalty. Further, we clarify the required assumptions on the elements of matrix $\bm{A}$ in each stage of simplification, which is a standard AMP derivation based on earlier studies \cite{Krzakala2012} for specific priors, but here we generalize the details to general priors/penalties/regularizations.

\subsection*{Stage 1: General form of belief propagation}

The coupled integral equations (\ref{m_mu_i}) (\ref{m_i_mu}) for the messages are too complicated to be of any practical use. However, in the large $N,~M$ limit, when the matrix elements $A_{\mu i}$ scale as $1/\sqrt{M}$ (or $1/\sqrt{N}$), these equations can be simplified.
We derive algebraic equations corresponding to
(\ref{m_mu_i}) and (\ref{m_i_mu}) by using sets of $a_{i\to\mu}$ and $\nu_{i \to \mu}$, which are the means and variances of $x_i$ under the posterior information message distributions defined in \eref{a_imu} and \eref{nu_imu}.
For this purpose, we define  $u_{\mu}\equiv\left(\bm{A}\bm{x}\right)_{\mu}$ and insert the identity
\begin{eqnarray}
1&=&\int\textrm{d}u_{\mu} \delta\left(u_{\mu}-\left(\bm{A}\bm{x}\right)_\mu\right)\nonumber\\
 &=&\int\textrm{d}u_{\mu} \frac{1}{2\pi}\int\textrm{d}\hat{u}_{\mu}\exp{\Biggl\{-i\hat{u}_{\mu}\left(u_{\mu}-\sum_{i=1}^{N}A_{\mu i}x_i\right)\Biggr\}}
\end{eqnarray}
into (\ref{m_mu_i}), which yields
\begin{eqnarray}
m_{\mu\rightarrow i}\left(x_i\right)
=\frac{1}{2\pi Z_{\mu\rightarrow i}} \int\textrm{d}u_{\mu}P\left(y_{\mu}|u_{\mu}\right)
 \int\textrm{d}\hat{u}_{\mu}\exp{\Biggl\{-i\hat{u}_{\mu}\left(u_{\mu}-A_{\mu i}x_i\right)\Biggr\}}\nonumber\\
\times\prod_{j \neq i}\Biggl\{\int\textrm{d}x_{j}m_{j\rightarrow \mu}\left( x_j\right)\exp{\Bigl\{ i \hat{u}_{\mu}A_{\mu j}x_j\Bigr\}} \Biggr\}.
\label{m_mu_i_1}
\end{eqnarray}
We truncate the Taylor series of $\exp\{ i \hat{u}_{\mu}A_{\mu j}x_j\}$ for $j\neq i$ up to the second order of $A_{\mu j}$.
By integrating $\int {\rm d}x_j m_{j\to \mu}(x_j) \left (\ldots \right )$ for $j \ne i$
and carrying out the resulting Gaussian integral of $\hat{u}_{\mu}$,
we obtain
\begin{eqnarray}
m_{\mu\rightarrow i}\left(x_i\right)
&=&\frac{1}{Z_{\mu\rightarrow i}
\sqrt{2\pi
(V_\mu -A_{\mu i}^2 \nu_{i \to \mu})
}}\int\textrm{d}u_{\mu}P\left(y_{\mu}|u_{\mu}\right)\nonumber\\
&&\times\exp{\Biggl\{-\frac{\left( u_{\mu}-\omega_\mu -A_{\mu i}(x_i -a_{i\rightarrow\mu})\right)^2}
{2(V_\mu -A_{\mu i}^2 \nu_{i \to \mu})}\Biggr\}},
\label{m_mu_i_2}
\end{eqnarray}
where $\omega_{\mu}$ and $V_{\mu}$ are defined in \eref{omega_df} and \eref{Vmu_df}, respectively.
We again truncate the
Taylor series of the exponential in (\ref{m_mu_i_2}) up to the second order on the basis of
the smallness of $A_{\mu i}$.
Consequently, a parameterized expression of $m_{\mu\rightarrow i}\left(x_i\right)$ is derived as
\begin{eqnarray}
m_{\mu\rightarrow i}\left(x_i\right)
=\frac{1}{\tilde{Z}_{\mu\rightarrow i}} \exp{\Biggl\{ -\frac{{\cal{A}}_{\mu\rightarrow i}}{2}x_{i}^2 +{\cal{B}}_{\mu \rightarrow i} x_i \Biggr\}},
\label{m_mu_i_3}
\end{eqnarray}
where $\tilde{Z}_{\mu\rightarrow i}=\sqrt{\frac{2\pi}{ {\cal A}_{\mu\rightarrow i}}}e^{\frac{{\cal{B}}^{2}_{\mu\rightarrow i}}{2{\cal{A}}_{\mu\rightarrow i}}}$.

The parameters ${\cal{A}}_{\mu \rightarrow i}$ and ${\cal{B}}_{\mu\rightarrow i}$ are evaluated as
\begin{eqnarray}
{\cal{A}}_{\mu\rightarrow i}=(g_{\rm out}^{\prime})_{\mu}A_{\mu i}^2\label{A}\\
{\cal{B}}_{\mu\rightarrow i}=(g_{\rm out})_{\mu}A_{\mu i}+(g_{\rm out}^{\prime})_{\mu}A_{\mu i}^2 a_{i\rightarrow \mu}\label{B}
\end{eqnarray}
using
 \begin{eqnarray}
(g_{\rm out})_{\mu} &\equiv& \frac{\partial }{\partial \omega_\mu}
\log \left (\int {\rm d} u_\mu P(y_\mu|u_\mu)\exp \left (-\frac{(u_\mu-\omega_\mu)^2}{2(V_\mu -A_{\mu i}^2 \nu_{i \to \mu})} \right ) \right )
\label{g_out_BP} \\
(g^\prime_{\rm out})_{\mu} &\equiv & -\frac{\partial^2 }{\partial \omega_\mu^2}
\log \left (\int {\rm d} u_\mu P(y_\mu|u_\mu)\exp \left (-\frac{(u_\mu-\omega_\mu)^2}{2(V_\mu -A_{\mu i}^2 \nu_{i \to \mu})} \right ) \right ).
\label{g_out_p_BP}
\end{eqnarray}
The relevant derivations are given in the appendix of \cite{Xu2014}.
Equations (\ref{A}) and (\ref{B}) act as algebraic expressions of (\ref{m_mu_i}). Note that the form of $g_{\rm out}$ and $g^\prime_{\rm out}$ only depends on the output distribution $P(y_\mu|u_\mu)$. We will give the specific expressions of $g_{\rm out}$ and $g^\prime_{\rm out}$ for linear regression later.

A similar expression for (\ref{m_i_mu}) is obtained by substituting the last expression of (\ref{m_mu_i_3})
into (\ref{m_i_mu}), which leads to
\begin{eqnarray}
m_{i\rightarrow\mu}(x_i)=\frac{1}{\tilde{Z}_{i\rightarrow\mu}}P_r(x_i)
 e^{-(x_{i}^2 /2)\sum\limits_{\gamma\neq\mu}{\cal{A}}_{\gamma\rightarrow i} +x_i\sum\limits_{\gamma\neq\mu}{\cal{B}}_{\gamma\rightarrow i} },
 \label{m_i_mu_alge}
\end{eqnarray}
where $\tilde{Z}_{i\rightarrow\mu}$ is a normalization constant.
The expression of equation (\ref{m_i_mu_alge}) indicates that $\prod_{\gamma \neq \mu}m_{\gamma\rightarrow i}\left(x_i\right)$ in (\ref{m_i_mu}) is expressed as a Gaussian distribution with
mean $R_{i\to\mu}$ and
variance $\Sigma_{i\to\mu}^2$,
given by
\begin{eqnarray}
R_{i\to\mu}&=\frac{\sum_{\gamma\neq\mu}{\cal{B}}_{\gamma\rightarrow i}}{\sum_{\gamma\neq\mu}{\cal{A}}_{\gamma\rightarrow i}},\label{df:R_imu}\\
\Sigma_{i\to\mu}^2&=\left(\sum_{\gamma\neq\mu}{\cal{A}}_{\gamma\rightarrow i}\right)^{-1}.
\end{eqnarray}

By referring to the definitions of $a_{i\rightarrow\mu}$ and $\nu_{i\rightarrow\mu}$ in (\ref{a_imu}) and (\ref{nu_imu}), we provide the closed form of BP update as
\begin{eqnarray}
a_{i\rightarrow\mu}=f_{a}\left(\Sigma_{i\to\mu}^2,R_{i\to\mu}\right)  \label{VstepA},\\
\nu_{i\rightarrow\mu}=f_{c}\left( \Sigma_{i\to\mu}^2,R_{i\to\mu}\right) \label{VstepNu}.
\end{eqnarray}
The moments of $m_i(x_i)$
are evaluated by adding the $\mu$-dependent part to (\ref{VstepA}) and (\ref{VstepNu}) as
\begin{eqnarray}
a_i=f_a (\Sigma_i^2,R_i) , \label{a}\\
\nu_i=f_c(\Sigma_i^2, R_i), \label{nu}
\end{eqnarray}
where
\begin{eqnarray}
\Sigma_i^2&=(\sum_{\mu}{\cal{A}}_{\mu\rightarrow i})^{-1}\\
R_i&=\frac{\sum_{\mu}{\cal{B}}_{\mu\rightarrow i}}{\sum_{\mu}{\cal{A}}_{\mu\rightarrow i}}\label{eq:Ri}.
\end{eqnarray}

Equations (\ref{a_imu}) and (\ref{nu_imu}) together with (\ref{A}), (\ref{B}), and (\ref{m_i_mu_alge}) lead to closed iterative message passing equations. These equations can be used for any data vector $\bm{y}$ and matrix $\bm{A}$.
We consider the case in which the matrix $\bm{A}$ is not sparse ($\sum_{i}A_{\mu i}x_i$ consists of order $N$ nonzero terms), and each element of the matrix scales as ${{O}}(1/\sqrt{M})$ (or ${{O}}(1/\sqrt{N})$).
Based on this fact, the use of means and variances instead of the canonical BP messages is exact in the large $N$ limit.

\subsection*{Stage 2: TAP form of the general message passing algorithm}
BP updates $2M\times N$ messages using (\ref{A}), (\ref{B}), (\ref{VstepA}), and
(\ref{VstepNu}) for $i=1,\cdots, N$ and $\mu=1,\cdots, M$ in each iteration.
The computational cost of this procedure is $O(M^2 \times N + M \times N^2)$ per iteration,
and limits the practical applicability of BP to systems of relatively small size.
In fact, it is possible to rewrite the BP equations in terms of $M+N$ messages instead of $2M \times N$ messages under the assumption that matrix $\bm{A}$ is not sparse and all its elements scale as $O(1/\sqrt{M})$ (or $O(1/\sqrt{N})$). In statistical physics, 
BP with reduced messages corresponds to the Thouless--Anderson--Palmer equations (TAP) \cite{Thouless1977} in the study of spin glasses. For large $N$, the TAP form is equivalent to the BP equations, and it is called the AMP algorithm \cite{Donoho2009} in compressed sensing. This form will result in a significant reduction of computational complexity to $O(M\times N)$ per iteration, which enhances the practical applicability of message passing.

First, we ignore $A_{\mu i}^2 \nu_{i \to \mu}$ in (\ref{m_mu_i_2}) for a sufficiently large system, because $A_{\mu i}^2$ vanishes as
$O(M^{-1})$ while $\nu_{i \to \mu} \sim O(1)$. Then, we obtain
 \begin{eqnarray}
(g_{\rm out})_{\mu} &\equiv& \frac{\partial }{\partial \omega_\mu}
\log \left (\int {\rm d} u_\mu P(y_\mu|u_\mu)\exp \left (-\frac{(u_\mu-\omega_\mu)^2}{2V_\mu} \right ) \right ),
\label{g_out_GAMP} \\
(g^\prime_{\rm out})_{\mu} &\equiv & -\frac{\partial^2 }{\partial \omega_\mu^2}
\log \left (\int {\rm d} u_\mu P(y_\mu|u_\mu)\exp \left (-\frac{(u_\mu-\omega_\mu)^2}{2V_\mu} \right ) \right ),
\label{g_out_p_GAMP}
\end{eqnarray}
which can be used in $\Sigma_i^2$ and $R_i$:
\begin{eqnarray}
\Sigma_{i}^{2}=\left( \sum_{\mu}(g_{\rm out}^{\prime})_{\mu}A_{\mu i}^{2} \right)^{-1}, \label{Sigma}\\
R_i=a_{i}+\left( \sum_{\mu}(g_{\rm out})_{\mu}A_{\mu i}\right)\Sigma_{i}^{2}.\label{R}
\end{eqnarray}

In the large $N$ limit, it is clear from (\ref{VstepA}) and (\ref{VstepNu}) that the messages $a_{i\rightarrow\mu}$ and $\nu_{i\rightarrow\mu}$ are nearly independent of $\mu$. However, one needs to
be careful to keep the correcting terms, which are referred to as the ``Onsager reaction terms'' \cite{Thouless1977,Shiino1992} in the spin glass literature.
We express $a_{i\to \mu}$ by applying Taylor's expansion to (\ref{VstepA}) around $R_i$ as
\begin{eqnarray}
a_{i\rightarrow\mu}
&=& f_{a}\left( \frac{1}{\sum_{\gamma}{\cal{A}}_{\gamma\rightarrow i}-{\cal{A}}_{\mu\rightarrow i}}, \frac{\sum_{\gamma}{\cal{B}}_{\gamma\rightarrow i}-{\cal{B}}_{\mu\rightarrow i}}{\sum_{\gamma}{\cal{A}}_{\gamma\rightarrow i}-{\cal{A}}_{\mu\rightarrow i}}\right)\nonumber\\
&\simeq & a_i
+\frac{\partial f_{a}(\Sigma^2_i,R_i)}{\partial R_i} (-{\cal{B}}_{\mu\rightarrow i}\Sigma_i^2)+ O(N^{-1}),
\label{TAP}
\end{eqnarray}
where ${\cal{B}}_{\mu\rightarrow i}\sim O(N^{-1/2})$  and
$\sum_{\gamma}{\cal{A}}_{\gamma\rightarrow i}-{\cal{A}}_{\mu\rightarrow i}$ is approximated as $\sum_{\gamma}{\cal{A}}_{\gamma\rightarrow i}=\Sigma_i^{-2}$
because of the smallness of ${\cal{A}}_{\mu\rightarrow i}\propto A_{\mu i}^2 \sim O(M^{-1})$.
According to definition (\ref{omega_df}), multiplying (\ref{TAP}) by $A_{\mu i}$  and summing  the resultant expressions over $i$
yields
\begin{eqnarray}
\omega_{\mu}
= \sum_{i} A_{\mu i}a_i -(g_{\rm out})_{\mu}V_\mu, \label{omega}
\end{eqnarray}
where we have used $\nu_i =f_c(\Sigma_i^2,R_i)=\Sigma_i^2 \frac{\partial f_{a}(\Sigma_i^2,R_i)}{\partial R_i}$. In addition, \eref{TAP} is used in the derivation of \eref{R} to replace $a_{i\rightarrow\mu}$ with $a_i$.

The computation of $V_{\mu}$ is similar. Since $\nu_{i\rightarrow\mu}$ is multiplied by $A_{\mu i}^2$, all the correction terms are negligible in the $N\rightarrow\infty$ limit. Therefore, we have
\begin{equation}
V_{\mu}=\sum_{i}A_{\mu i}^{2}\nu_{i}.
\label{eq:V_mu_expanded}
\end{equation}

The general TAP form of the message passing or general approximated message passing algorithm is summarized in Figure \ref{GAMP}.

\begin{figure}
\renewcommand{\thepseudocode}{\arabic{pseudocode}}
\setcounter{pseudocode}{1}
\begin{pseudocode}[ruled]
{General Approximate Message Passing}
{\mathbf{a}^*, \mathbf{\nu}^*, \mathbf{\omega}^*}

1)\ \mbox{\bf Initialization}:\\
 \hspace{15pt}\bm{a}\textrm{ seed}: \hspace{65pt} \bm{a}_{0} \GETS \bm{a}^*\\
 \hspace{15pt}\bm{\nu}\textrm{ seed}: \hspace{65pt} \bm{\nu}_0 \GETS \bm{\nu}^* \\
 \hspace{15pt}\bm{\omega}\textrm{ seed}: \hspace{65pt} \bm{\omega}_0 \GETS \bm{\omega}^* \\
 \hspace{15pt}\textrm{Counter}: \hspace{62pt} t\GETS 0\\

2)\ \mbox{\bf Counter increase}:\\
 \hspace{30pt}t \GETS t+1\\

3)\ \mbox{\bf Mean of variances of posterior information message distributions}:\\
 \hspace{30pt}{V_{\mu}}^{(t)}\GETS  ({\sum}_{i} A_{\mu i}^{2}{\nu}^{(t-1)})\\

4)\ \mbox{\bf Self-feedback cancellation}:\\
 \hspace{30pt}
 {\omega}_{\mu}^{(t)} \GETS {\sum}_{i}{A_{\mu i}}{a_{i}}^{(t-1)}-{V}_{\mu}^{(t)}g_{\rm out}({\omega}_{\mu}^{(t-1)}, {V}_{\mu}^{(t)})\\

5)\ \mbox{\bf Variances of output information message distributions}:\\
 \hspace{30pt}(\Sigma_{i}^{2})^{(t)} \GETS \left({\sum}_{\mu} A_{\mu i}^{2} g_{\rm out}^{\prime}({\omega}_{\mu}^{(t)}, {V}_{\mu}^{(t)})\right)^{-1}\\

6)\ \mbox{\bf Average of output information message distributions}:\\
 \hspace{30pt}{R}_{i}^{(t)}\GETS {a}_{i}^{(t-1)}+\left({\sum}_{\mu}{A}_{\mu i}g_{\rm out}({\omega}_{\mu}^{(t)}, {V}_{\mu}^{(t)})\right)(\Sigma_i^2)^{(t)}\\

7)\ \mbox{\bf Posterior mean}:\\
 \hspace{30pt}{a}_{i}^{(t)}\GETS f_a ((\Sigma_{i}^{2})^{(t)}, {R}_{i}^{(t)})\\

8)\ \mbox{\bf Posterior variance}:\\
 \hspace{30pt}{\nu}_{i}^{(t)}\GETS f_c ((\Sigma_{i}^{2})^{(t)}, {R}_{i}^{(t)})\\

9)\ \bf{Iteration}: \mbox{Repeat from step 2 until convergence.}
\end{pseudocode}
\protect
\caption{GAMP algorithm with the assumption that matrix $\bm{A}$ is not sparse and all its elements scale as $O(1/\sqrt{M})$ (or $O(1/\sqrt{N})$).
The convergent vectors of $\bm{a}^{(t)}$, $\bm{\nu}^{(t)}$,
and $\bm{\omega}^{(t)}$ obtained in the previous loop are denoted by $\bm{a}^*$, $\bm{\nu}^*$, and $\bm{\omega}^*$, respectively.}
\label{GAMP}
\end{figure}

\subsection*{Stage 3: Further simplification for basis matrix with random entries}
\label{sec:GAMP_iidA}
For special cases of random matrix $\bm{A}$, the TAP equations can be simplified further. For homogeneous matrix $\bm{A}$ with i.i.d random entries of zero mean and variance $1/M$ (the distribution can be anything as long as the mean and variance are fixed), the simplification can be understood as follows.
Define ${V}$ as the average of $V_{\mu}$ with respect to different realisations of the matrix $\bm{A}$.
For high dimensional cases where $N, M\rightarrow\infty$, the Onsager terms are negligible in $V_{\mu}= \sum_{i}A_{\mu i}^2\nu_{i\rightarrow \mu}$ and we have the approximated uncorrelated expression \eref{eq:V_mu_expanded}. Therefore, effectively, we could replace $\{A_{\mu j}^2\}$ in $V_{\mu}= \sum_{i}A_{\mu i}^2\nu_{i}$
with its expectation
$M^{-1}$ from the law of large numbers.
This replacement makes $V_{\mu}$ for any $\mu$ equal to the average as
\begin{equation}
V\equiv\sum_{i=1}^{N}\overline{A_{\mu i}^2}\nu_{i}=\frac{1}{M}\sum_{i=1}^{N}\nu_{i},
\label{V}
\end{equation}
where $\overline{\cdots}$ denotes the average over $A_{\mu i}$.
The same argument can be repeated for all the terms that contain $A_{\mu i}$. For sufficiently large $N$, $\Sigma^{2}_{i}$ typically converges to a constant
denoted by $\Sigma^2$, independent of index $i$. This removal of site dependence, in conjunction with (\ref{A}) and (\ref{B}), yields
\begin{eqnarray}
\Sigma^2=\left( \frac{1}{M}\sum_{\mu}g_{\rm out}^{\prime}(\omega_{\mu},V) \right)^{-1}, \label{Sigma_iidA}\\
R_i=\left( \sum_{\mu}g_{\rm out}(\omega_{\mu},V)A_{\mu i}\right)\Sigma^2 +a_i.\label{R_iidA}
\end{eqnarray}
By iterating (\ref{V}), (\ref{Sigma_iidA}), (\ref{R_iidA}), and
\begin{eqnarray}
\omega_{\mu}
&= \sum_{i} A_{\mu i}a_i -g_{\rm out}(\omega_{\mu},V)V, \label{omega_iidA}\\
a_{i}&=f_{a}(\Sigma^{2}, R_i),\\
\nu_{i}&=f_{c}(\Sigma^{2},R_i),
\end{eqnarray}
we can solve the equations with only $2(M+N+1)$ variables involved.

The most time-consuming parts of GAMP are the matrix-vector multiplications
$\sum_{\mu}(g_{\rm out})_{\mu}A_{\mu i}$ in (\ref{R}) and
$\sum_{i} A_{\mu i} a_i$ in (\ref{omega}); hence, the computational complexity is $O(NM)$ per iteration.

In this paper, we consider the case in which matrix $\bm{A}$ contains i.i.d random variables from a Gaussian distribution with zero mean and variance $1/M$. We can employ the simplified version of the GAMP algorithm derived above.

\section{Replica analysis for SCAD
 regularization}
\label{sec:replica_SCAD}

By assuming that $n$ is an integer,
we can express the expectation of $Z_\beta^n$
in \eref{eq:replica}
using $n$-replicated systems:
\begin{eqnarray}
\nonumber
E_{\bm{y},\bm{A}}[Z_{\beta}^n(\bm{y},\bm{A})]&=\int d\bm{A}d\bm{y}P_{A,y}(\bm{A},\bm{y})\int
 d\bm{x}^{(1)}\cdots d\bm{x}^{(n)}\\
&\times\exp\Big[\sum_{b=1}^n\Big\{-\frac{\beta}{2}||\bm{y}-\bm{Ax}^{(b)}||_2^2-\beta J_{\lambda,a}(\bm{x}^{(b)})\Big\}\Big].
\label{eq:E_replica}
\end{eqnarray}
The microscopic states of $\{\bm{x}^{(b)}\}$
are characterized by the macroscopic quantities
\begin{eqnarray}
q^{(bc)}&=\frac{1}{N}\sum_ix_i^{(b)}x_i^{(c)},
\end{eqnarray}
which are defined for all combinations of
replica indices $b$ and $c$ $(b\geq c)$.
By introducing the identity
\begin{eqnarray}
1=\int dq^{(bc)}
\delta\left(q^{(bc)}-\frac{1}{M}\sum_ix_i^{(b)}x_i^{(c)}\right)
\label{eq:subshell}
\end{eqnarray}
for all combinations of $(b,c)$, the following expression is obtained
after integration with respect to $\bm{A}$:
\begin{eqnarray}
E_{\bm{y},\bm{A}}[Z_{\beta}^n(\bm{y},\bm{A})]&=\int d{\cal Q}{\cal
 S}({\cal Q}){\cal I}({\cal Q}),
 \label{eq:Z_n}
\end{eqnarray}
where the functions ${\cal S}({\cal Q})$ and ${\cal I}({\cal Q})$ are given by
\begin{eqnarray}
\nonumber
{\cal S}({\cal Q})&=\int d\hat{\cal Q}\{d\bm{x}^{(b)}\}\exp\Big\{-M{\sum_{b\leq c}q^{(bc)}\hat{q}^{(bc)}}+\sum_{b\leq
  c}\sum_i\hat{q}^{(bc)}x_i^{(b)}x_i^{(c)}\\
&-\beta \sum_{b=1}^nJ_{\lambda,a}(\bm{x}^{(b)})\Big\},\\
{\cal I}({\cal Q})&=\int \{d\bm{u}^{(b)}\}P_u(\{\bm{u}^{(b)}\}|{\cal Q})\!\int\! d\bm{y}P_y(\bm{y})\exp\!\Big\{\!-\frac{\beta}{2}\sum_{b=1}^n||\bm{y}-\bm{u}^{(b)}||^2_2\!\Big\}.
\end{eqnarray}
The $n\times n$ matrices ${\cal Q}$ and $\hat{\cal Q}$
are the matrix representations of $\{q^{(bc)}\}$ and
$\{\hat{q}^{(bc)}\}$, respectively,
where $\{\hat{q}^{(bc)}\}$ are conjugate variables for the
representation of the delta function in \eref{eq:subshell}.
Each component of $\bm{u}^{(b)}\in\mathbb{R}^M$, denoted by $u_\mu^{(b)}$,
is statistically equivalent to $\sum_iA_{\mu i}x_i^{(b)}$.
The probability distribution of $\{\bm{u}^{(b)}\}$
is given by the product of the Gaussian distribution with
respect to the vector $\tilde{\bm{u}}_\mu=\{u_\mu^{(1)},\cdots,u_\mu^{(n)}\}$,
such as \cite{Kabashima2009}
\begin{eqnarray}
P_u(\{\bm{u}^{(b)}\}|{\cal Q})=\prod_{\mu=1}^M\frac{1}{\sqrt{(2\pi)^n|{\cal
 Q|}}}\exp\Big(-\frac{1}{2}\tilde{\bm{u}}_\mu^{\rm T}{\cal Q}^{-1}\tilde{\bm{u}}_\mu\Big).
\end{eqnarray}

To obtain an analytic expression with respect to $n\in\mathbb{R}$
and taking the limit as $n\to 0$,
we restrict the candidates for the dominant saddle point
to those of the RS form as
\begin{eqnarray}
(q^{(bc)},\hat{q}^{(bc)})=\left\{
\begin{array}{ll}
(Q,-\tilde{Q}\slash 2) & (b=c) \\
(q,\tilde{q}) & (b\neq c).
\end{array}\right.\label{eq:q_RS}
\end{eqnarray}
Under the RS assumption \eref{eq:q_RS}, we obtain
\begin{eqnarray}
\nonumber
{\cal S}(Q,q)&=\int d\tilde{Q}d\tilde{q}\exp\Big[M\Big\{\frac{-nQ\tilde{Q}+n(n-1)q\tilde{q}}{2}
\\
&+\log\int Dz\left(\int dx\exp\left(\frac{\tilde{Q}}{2}x^2+\sqrt{\tilde{q}}z-\beta J_{\lambda,a}(x)\right)\right)^n\Big\}\Big]\label{eq:S_RS}\\
{\cal I}(Q,q)&=
\left[\int \!\! DwdyP_y(y)\left\{\!\!\int\!\! Dv\exp\!\left(\!-\frac{\beta}{2}(y\!-\!\sqrt{Q\!-\!q}v\!-\!\sqrt{q}w)^2\!\right)\!\right\}^n\right]^M
\label{eq:I_RS}
\end{eqnarray}
For $\beta\to\infty$, the RS order parameters
scale to keep $\beta(Q-q)=\chi$,
$\beta^{-1}(\tilde{Q}+\tilde{q})=\hat{Q}$,
and $\beta^{-2}\tilde{q}=\hat{\chi}$
of the order of unity.
The integral with respect to $x$
in \eref{eq:S_RS}
is processed by the saddle point method at $\beta\to\infty$,
which corresponds to the single-body problem of the estimate \eref{eq:single_body_SCAD}.
By substituting \eref{eq:Z_n}, \eref{eq:S_RS}, and \eref{eq:I_RS} into \eref{eq:replica},
we obtain the RS free energy density \eref{eq:free_energy}.

The free energy density corresponds to the minimum value of the energy.
Therefore, one can derive the
representation error \eref{eq:def_err}
by discarding the
expectation of the regularization term from the free energy density as
\begin{eqnarray}
{\rm err}=2\left(f-\frac{1}{M}E_{\bm{y},\bm{A}}[J_{\lambda,a}(\hat{\bm{x}}(\bm{y},\bm{A})]\right).
\label{eq:err_def}
\end{eqnarray}
As mentioned in Section \ref{sec:replica},
$x^*$ is statistically equivalent to the solution of the original problem.
Hence, the expectation value of the regularization term is
derived as
\begin{eqnarray}
\lim_{N\to\infty}\frac{1}{M}E_{\bm{y},\bm{A}}[J_{\lambda,a}(\hat{\bm{x}}(\bm{y},\bm{A}))]=\frac{1}{\alpha}\int Dz J_{\lambda,a}(x^*(z;\hat{Q},\hat{\chi})).
\label{eq:reg_ave}
\end{eqnarray}
By substituting the expression of \eref{eq:reg_ave} under RS assumption into \eref{eq:err_def},
we obtain the relationship $\hat{\chi}={\rm err}$.

\section{Derivation of de Almeida--Thouless condition}
\label{sec:replica_AT}

The local stability of the RS solution against
symmetry breaking perturbation
is examined by
constructing the one-step replica symmetry breaking (1RSB) solution.
We introduce the 1RSB assumption with respect to the saddle point as
\begin{eqnarray}
(q^{(bc)},\hat{q}^{(bc)})=\left\{
\begin{array}{ll}
(Q,-\tilde{Q}\slash 2) & (b=c) \\
(q_1,\tilde{q}_1) & (b\neq c,~B(b)=B(c))\\
(q_0,\tilde{q}_0) & (B(b)\neq B(c))
\end{array}\right.,
\end{eqnarray}
where $n$ replicas are separated into $\tilde{l}$ blocks
that contain $n\slash \tilde{l}$ replicas.
The index of the block
that includes the $b$-th replica is denoted by $B(b)$.
At $\beta\to\infty$,
these 1RSB order parameters
scale as
$\beta(Q-q_1)=\chi$,
$\beta^{-1}(\tilde{Q}+\tilde{q}_1)=\hat{Q}$,
$\beta^{-2}\tilde{q}_1=\hat{\chi}_1$,
$\beta^{-2}\tilde{q}_0=\hat{\chi}_0$,
and $\tilde{l}=l\slash \beta$.
Following the calculation
explained in \ref{sec:replica_SCAD},
the free energy density under
the 1RSB assumption is derived as
\begin{eqnarray}
\nonumber
f&=\frac{\alpha(Q+\sigma_y^2)}{2\{1+\chi+l(Q-q_0)\}}-\frac{\alpha(\hat{Q}Q-\hat{\chi}_1\chi)}{2}+\frac{\alpha l(\hat{\chi}_1Q-\hat{\chi}_0q_0)}{2}\\
&-\frac{1}{l}\int Dz\log \Xi+\frac{\alpha}{2l}\log\left(1+\frac{l(Q-q_0)}{1+\chi}\right),
\end{eqnarray}
where
\begin{eqnarray}
&\Xi =\int Dw \exp(lf_\xi^*)\\
&f_\xi^*=\min_x f_\xi(x)\label{eq:f_xi_RSB}
\\
&f_\xi(x) = \frac{\hat{Q}}{2}x^2-(\sqrt{\hat{\chi_1}}z+\sqrt{\hat{\chi_1}-\hat{\chi_0}}w)x+J_{\lambda,a}(x).
\end{eqnarray}
We denote the minimizer of \eref{eq:f_xi_RSB} as $x^*_{\rm RSB}$.
The saddle point equations are
given by
\begin{eqnarray}
\hat{Q}&=\frac{1}{1+\chi}\\
\hat{\chi}_1&=\frac{Q-q_0}{(1+\chi)(1+\chi+l(Q-q_0))}+\hat{\chi}_0\label{eq:hchi_1}\\
\hat{\chi}_0&=\frac{Q+\sigma_y^2}{(1+\chi+l(Q-q_0))^2}\label{eq:hchi_0}\\
Q&=\frac{1}{\alpha}\int Dz\langle {x^{*2}_{\rm RSB}}\rangle_w\\
\chi&=\frac{1}{\alpha}\int Dz\left<\frac{\partial x^*_{\rm RSB}}{\partial(\sqrt{\hat{\chi}_1-\hat{\chi}_0}w)}\right>_w\\
q_0&=\frac{1}{\alpha}\int Dz\langle x^*_{\rm RSB}\rangle_w^2,
\end{eqnarray}
where
\begin{eqnarray}
\langle g(w)\rangle_w=\frac{\int Dw \exp(lf_\xi^*) g(w)}{\int Dw \exp(lf_\xi^*)}
\label{eq:w_ave_def}
\end{eqnarray}
with an arbitrary function $g(w)$.
When $q_0= Q$ and $\hat{\chi}_0=\hat{\chi}_1$,
the 1RSB solution is reduced to
the RS solution. Hence,
the stability of the RS solution
is examined by the stability analysis of the
1RSB solution of $Q=q_0$
and $\hat{\chi}_1=\hat{\chi}_0$.
We define
$\Delta=Q-q_0$ and
$\hat{\Delta}=\hat{\chi}_1-\hat{\chi}_0$,
and we apply Taylor expansion to them
by assuming that they
are sufficiently small.
From \eref{eq:hchi_1} and \eref{eq:hchi_0},
we obtain
\begin{eqnarray}
\hat{\Delta}=\frac{1}{(1+\chi)^2}\Delta+O(\Delta^2).
\label{eq:dhD}
\end{eqnarray}
For the expansion of $\Delta$
around $\hat{\Delta}=0$,
the saddle point equations for
finite $\beta$
are useful. They are
obtained by replacing $x^*_{\rm RSB}$ with
\begin{eqnarray}
\langle x\rangle&=\frac{\int dx x\exp(\beta f_\xi)}{\int dx\exp(\beta f_\xi)},
\end{eqnarray}
and $\langle g(w)\rangle_w$
for finite $\beta$
is defined by
\begin{eqnarray}
\langle g(w)\rangle_w&=\frac{\int Dw\left\{\int dx\exp(\beta f_\xi)\right\}^{\tilde{l}}g(w)}{\int Dw\left\{\int dx\exp(\beta f_\xi)\right\}^{\tilde{l}}}
\end{eqnarray}
because this expression
depends on $\hat{\Delta}$ only
through $f_\xi$.
By differentiating $Q-q_0$
by $\hat{\Delta}$
at $\hat{\Delta}=0$,
we obtain
\begin{eqnarray}
\frac{\partial \Delta}{\partial\hat{\Delta}}\Big|_{\hat{\Delta}= 0}=\frac{\beta^2}{\alpha}\int Dz
\left\{\langle x^2\rangle^2-2\langle x^2\rangle\langle x\rangle^2+\langle x\rangle^4\right\},
\label{eq:dD}
\end{eqnarray}
which is equivalent to
\begin{eqnarray}
\frac{\partial\Delta}{\partial\hat{\Delta}}\Big|_{\hat{\Delta}=0}=\frac{1}{\alpha}\int Dz\left(\frac{\partial x^*}{\partial \sqrt{\hat{\chi}}z}\right)^2
\label{eq:dD2}
\end{eqnarray}
at $\beta\to\infty$.
By substituting \eref{eq:dD2}
into \eref{eq:dhD},
we get
\begin{eqnarray}
\hat{\Delta}=\frac{1}{\alpha(1+\chi)^2}\int Dz\left(\frac{\partial x^*}{\partial \sqrt{\hat{\chi}}z}\right)^2\hat{\Delta}+O(\hat{\Delta}^2),
\end{eqnarray}
which means that
the local stability of $\hat{\Delta}=0$ is lost when
\begin{eqnarray}
\frac{1}{\alpha(1+\chi)^2}\int Dz\left(\frac{\partial x^*}{\partial \sqrt{\hat{\chi}}z}\right)^2>1.
\end{eqnarray}
This is the de Almeida--Thouless condition for the current problem.

\section*{References}

\providecommand{\newblock}{}

\end{document}